%% file: bare_adv.tex
\documentclass[10pt,journal,compsoc]{IEEEtran}

\ifCLASSOPTIONcompsoc
\usepackage[utf8]{inputenc} 
\usepackage[T1]{fontenc}    
\usepackage{hyperref}       
\usepackage{url}            
\usepackage{booktabs}       
\usepackage{amsfonts}       
\usepackage{nicefrac}       
\usepackage{microtype}      
\usepackage{bm}
\usepackage{amsmath}
\usepackage{algorithmic}
\usepackage{algorithm}
\usepackage{multirow}
\usepackage{graphicx}
\usepackage{subfigure}
\usepackage{wrapfig}
\usepackage{amsthm}
\usepackage{multicol}
\usepackage{amssymb}
\usepackage{pdfpages}
\usepackage{makecell}

\usepackage{pifont}

\newenvironment{shrinkeq}[1]
{\bgroup
  \addtolength\abovedisplayshortskip{#1}
  \addtolength\abovedisplayskip{#1}
  \addtolength\belowdisplayshortskip{#1}
  \addtolength\belowdisplayskip{#1}}
{\egroup\ignorespacesafterend}

  \usepackage[nocompress]{cite}
\else
  \usepackage{cite}
\fi
\ifCLASSINFOpdf
\else
\fi

\usepackage{stfloats}
\vspace{5in}
\IEEEpubid{
\fbox{
\parbox{0.975\textwidth}{
\copyright 2022 IEEE. Personal use of this material is permitted. Permission from IEEE must be obtained for all other uses, in any current or future media, including\hfill
reprinting/republishing this material for advertising or promotional purposes, creating new collective works, for resale or redistribution to servers or lists, or reuse of\hfill
any copyrighted component of this work in other works. \vspace{-0.5mm}
}%
}}

\begin{document}
%

\title{Glance and Focus Networks \\for Dynamic Visual Recognition}

%
%
%
%


\author{Gao~Huang\IEEEauthorrefmark{1},~\IEEEmembership{Member,~IEEE},
        Yulin~Wang\IEEEauthorrefmark{1},
        Kangchen~Lv, 
        Haojun~Jiang, 
        Wenhui~Huang,
        Pengfei~Qi,
        and~Shiji~Song,~\IEEEmembership{Senior~Member,~IEEE}
\IEEEcompsocitemizethanks{
  \IEEEcompsocthanksitem G. Huang, Y. Wang, K. Lv, H. Jiang and S. Song are with the Department of Automation, BNRist, Tsinghua University, Beijing 100084, China. G. Huang is also with Beijing Academy of Artificial Intelligence (BAAI). Email: \{gaohuang, shijis\}@tsinghua.edu.cn, \{wang-yl19, lkc21, jhj20\}@mails.tsinghua.edu.cn. Corresponding author: Shiji Song.
  \IEEEcompsocthanksitem W. Huang and P. Qi are with the China Mobile Research Insitute, Beijing 100053, China. Email: \{huangwenhui, qipengfei\}@chinamobile.com.

}
}

%
%

\markboth{IEEE transactions on pattern analysis and machine intelligence}%
{Shell \MakeLowercase{\textit{et al.}}: Bare Advanced Demo of IEEEtran.cls for IEEE Computer Society Journals}
%



\IEEEtitleabstractindextext{%
\begin{abstract}

Spatial redundancy widely exists in visual recognition tasks, \emph{i.e.}, discriminative features in an image or video frame usually correspond to only a subset of pixels, while the remaining regions are irrelevant to the task at hand. Therefore, static models which process all the pixels with an equal amount of computation result in considerable redundancy in terms of time and space consumption. In this paper, we formulate the image recognition problem as a sequential coarse-to-fine feature learning process, mimicking the human visual system. Specifically, the proposed Glance and Focus Network (GFNet) first extracts a quick global representation of the input image at a low resolution scale, and then strategically attends to a series of salient (small) regions to learn finer features. The sequential process naturally facilitates adaptive inference at test time, as it can be terminated once the model is sufficiently confident about its prediction, avoiding further redundant computation. It is worth noting that the problem of locating discriminant regions in our model is formulated as a reinforcement learning task, thus requiring no additional manual annotations other than classification labels. GFNet is general and flexible as it is compatible with any off-the-shelf backbone models (such as MobileNets, EfficientNets and TSM), which can be conveniently deployed as the feature extractor. Extensive experiments on a variety of image classification and video recognition tasks and with various backbone models demonstrate the remarkable efficiency of our method. For example, it reduces the average latency of the highly efficient MobileNet-V3 on an iPhone XS Max by 1.3x without sacrificing accuracy. Code and pre-trained models are available at \url{https://github.com/blackfeather-wang/GFNet-Pytorch}.

\end{abstract}

}

\maketitle

\begingroup\renewcommand\thefootnote{\IEEEauthorrefmark{1}}
\footnotetext{\emph{Equal contribution.}}

\IEEEdisplaynontitleabstractindextext

%
\IEEEpeerreviewmaketitle

\input{introduction.tex}
\input{related.tex}
\input{method.tex}
\input{experiments.tex}

\input{conclusion.tex}

\ifCLASSOPTIONcompsoc
  \section*{Acknowledgments}
\else
  \section*{Acknowledgment}
\fi


This work is supported in part by the National Key R\&D Program of China under Grant 2019YFC1408703, the National Natural Science Foundation of China under Grants 61906106 and 62022048, the Tsinghua University-China Mobile Communications Group Co.,Ltd. Joint Institute and Guoqiang Institute of Tsinghua University. We also appreciate the donation of computational resources from Hangzhou High-Flyer AI Fundamental Research Co.,Ltd.

\ifCLASSOPTIONcaptionsoff
  \newpage
\fi



%



\bibliographystyle{IEEEtran}
\bibliography{IEEEabrv,IEEEtran}

%








\newpage
\twocolumn

\input{appendix.tex}

\end{document}

%% file: introduction.tex

\section{Introduction}

\IEEEPARstart{T}{he} availability of high resolution images or videos enabled deep learning algorithms to achieve super-human-level performance on many challenging vision tasks \cite{5206848,2014arXiv1409.1556S, He_2016_CVPR, xie2017aggregated, hu2018squeeze, 2019arXiv160806993H}. Recent works \cite{DBLP:conf/icml/TanL19, huang2019gpipe} show that the accuracy of deep models can be further boosted by scaling up the image resolution, \emph{e.g.}, to 480$\times$480 or even higher. Although acquiring high quality data with modern cameras is easy, performing visual recognition on them proves to be challenging in practice, due to the high computational cost and high memory footprint introduced by deep neural networks. In real-world applications like visual recognition on edge devices \cite{howard2017mobilenets}, content-based image search \cite{wan2014deep} or autonomous vehicles \cite{bojarski2016end}, computation translates into latency and power consumption, which should be minimized for both safety and economical reasons \cite{howard2017mobilenets, huang2018condensenet, sandler2018mobilenetv2}.

\begin{figure}[t]
    \begin{minipage}[t]{\linewidth}
    \centering
    \includegraphics[width=0.74\textwidth]{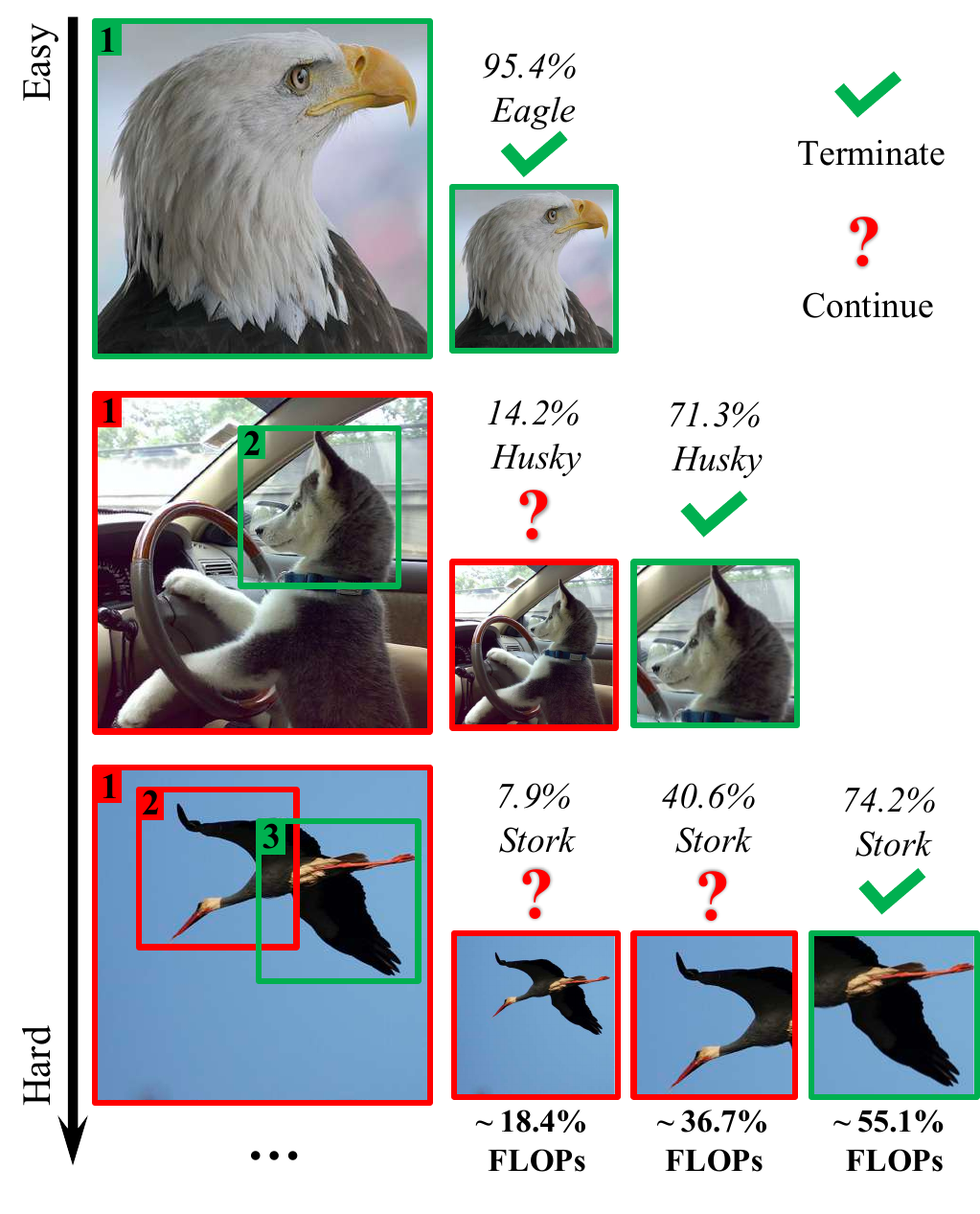}	
    \vskip -0.175in
    \caption{Examples for GFNet. ``FLOPs'' refers to the proportion of the computation required by GFNet (with 96$\times$96 image patches) versus processing the entire 224$\times$224 image.
    \label{fig:motivation}}  
    \end{minipage}
    \vskip -0.175in
\end{figure}

This paper aims to reduce the computational cost of high-resolution visual recognition from the perspective of \textit{spatial redundancy}. In fact, deep models are shown to be able to perform  object recognition accurately with only a few class-discriminative patches, such as the head of a dog or the wings of a bird \cite{mnih2014recurrent, fu2017look, chu2019spot}. These regions typically contribute to a small fraction of the whole image, and thus require much less computational resources to process. Therefore, if we can dynamically identify the ``class-discriminative'' regions of each individual image, and perform efficient inference only on these small input patches, then the spatial-wise computational redundancy can be significantly reduced without sacrificing accuracy. To implement this idea, we need to address two challenges: 1) how to efficiently identify class-discriminative regions (without resorting to additional human annotations); and 2) how to adaptively allocate computation to each individual image, given that the number/size of discriminative regions may differ across different inputs.

To address the aforementioned issues, we present a two-stage framework, named \emph{glance and focus}, which is motivated by the coarse-to-fine recognition procedure of human visual systems. Specifically, the region selection operation is formulated as a sequential decision process, where at each step our model processes a relatively small input, producing a classification prediction with a confidence score as well as a region proposal for the next step. Each step can be done efficiently due to the reduced input size. For example, the computational cost of inferring a $96\times 96$ image patch is only $18\%$ of that of processing the original $224\times 224$ input. The whole sequential process starts with processing the full image in a down-sampled scale (\emph{e.g.}, $96\times 96$), serving as the initial step. We call it the \emph{glance step}, at which the model produces a quick prediction with \emph{global} features. In practice, we find that a large portion of images with discriminative features can already be correctly classified with high confidence at the \emph{glance step}, which is inline with the observation in \cite{yang2020resolution}. When the \emph{glance step} fails to produce sufficiently high confidence
about its prediction, it will output a proposal (relative location within an image) of the most discriminative region for the subsequent step to process. As the proposed region is usually a small patch of the original image with full resolution, we call these subsequent steps the \emph{focus stage}. This stage proceeds progressively with iteratively localizing and processing the class-discriminative image regions, facilitating early termination in an adaptive manner, \emph{{i.e.},} the decision process can be interrupted dynamically conditioned on each input image. As shown in Figure \ref{fig:motivation}, our method allocates computation unevenly across different images at test time, leading to a significant improvement of the overall efficiency. We refer to our method as \textit{Glance and Focus Network (GFNet)}.

One notable feature of the proposed GFNet is that it is a general framework, wherein the classifier and the region proposal network are treated as two independent modules. Therefore, any existing backbone models, such as MobileNets \cite{howard2017mobilenets, sandler2018mobilenetv2, howard2019searching}, CondenseNets \cite{huang2018condensenet, yang2021condensenetv2}, ShuffleNets \cite{zhang2018shufflenet, ma2018shufflenet} and EfficientNets \cite{DBLP:conf/icml/TanL19}, can be deployed as our feature extractors. This differentiates our method from early recurrent attention methods \cite{mnih2014recurrent} which adopt pure recurrent models. In addition, we focus on improving the computational efficiency under the adaptive inference setting, while most existing works aim to improve accuracy with the fixed sequence length.

Besides its high computational efficiency, the proposed GFNet is appealing in several other aspects. For example, the memory consumption can be significantly reduced, and it is independent of the original image resolution as long as we fix the size of the focus region. Moreover, the computational cost of GFNet can be adjusted online without additional training (by simply adjusting the termination criterion). This enables GFNet to make full use of all available computational resources flexibly or achieve the required performance with minimal power consumption -- a practical requirement of many real-world applications such as search engines and mobile apps.

We empirically validates GFNet on image classification (ImageNet \cite{5206848}) and video recognition tasks (Something-Something V1\&V2 \cite{goyal2017something} and Jester \cite{materzynska2019jester}) with various backbone models (\emph{e.g.}, MobileNet-V3 \cite{howard2019searching}, RegNet \cite{radosavovic2020designing}, EfficientNet \cite{DBLP:conf/icml/TanL19} and TSM \cite{lin2019tsm}). Two practical settings, \emph{i.e.}, the budgeted batch classification setting \cite{huang2017multi}, where the test set comes with a given computational budget, and the anytime prediction setting \cite{grubb2012speedboost, huang2017multi}, where the network can be forced to output a prediction at any given point in time, are considered. 
We also benchmark the practical speed of GFNet on an iPhone XS Max and a NVIDIA 2080Ti GPU. Experimental results show that GFNet effectively improves the efficiency of state-of-the-art networks both theoretically and empirically. For example, when the MobileNets-V3 and ResNets are used as the backbone network, GFNet yields up to $1.4 \times $ and $3\times$ less \emph{Multiply-Add} operations compared to the original models when achieving the same level of accuracy, respectively. Notably, the actual speedup on the iPhone XS Max (measured by average latency) is $1.3\times $ and $2.9\times$, respectively.

Parts of the results in this paper were published originally in its conference version \cite{NeurIPS2020_7866}. However, this version extends our earlier work in several important aspects:
\begin{itemize}
    \item We improve the reinforcement learning algorithm for learning the patch selection policy by proposing a contrast reward function (Section \ref{sec:contrastive_reward}).
    \item We introduce a multi-scale GFNet (MS-GFNet), allowing adjusting the patch size dynamically conditioned on the inputs and the computational budgets (Sections \ref{sec:MS-GFNet}). Empirical results indicate that MS-GFNet not only improves the accuracy, but also yields a more flexible range of computational cost for online tuning.
    \item We extend the proposed GFNet to high-resolution or non object-centric image recognition and video recognition, and report encouraging results on widely used benchmarks (Sections \ref{sec:high_res}, \ref{sec:video_recognition}).
    \item We conduct experiments to verify that GFNet can also effectively accelerate the large-batch inference on GPU devices (Table \ref{tab:gpu_speed}), while the original version only considers the single-image inference on an iPhone.
    \item Additional analytical results are presented, including the comprehensive ablation studies and a discussion on training hyper-parameters (Sections \ref{sec:abl_study}, \ref{sec:hyper_sense}).
\end{itemize}

%% file: related.tex
\section{Related Work}
\subsection{Computationally efficient networks}
Modern deep learning models such as convolutional neural networks (CNNs) usually require a large number of computational resources. To this end, many research works focus on reducing the inference cost of the networks. A promising direction is to develop efficient network architectures, such as MobileNets \cite{howard2017mobilenets, sandler2018mobilenetv2, howard2019searching}, CondenseNets \cite{huang2018condensenet, yang2021condensenetv2}, ShuffleNets \cite{zhang2018shufflenet, ma2018shufflenet} and EfficientNet \cite{DBLP:conf/icml/TanL19}. Since deep networks typically have a considerable number of redundant weights \cite{frankle2018lottery}, some other approaches focus on pruning \cite{lecun1990optimal, li2017pruning, liu2017learning, liu2018rethinking} or quantizing the weights \cite{rastegari2016xnor, hubara2016binarized, jacob2018quantization}. Another technique is knowledge distillation \cite{hinton2015distilling}, which trains a small network to reproduce the prediction of a large model. Our method is orthogonal to the aforementioned approaches, and can be combined with them to further improve the efficiency.

A number of recent works improve the efficiency of deep models by \emph{adaptively} changing the architecture of the network \cite{han2021dynamic, xia2022vision}. For example, MSDNet \cite{huang2017multi} and its variants \cite{li2019improved, yang2020resolution} introduce a multi-scale architecture with multiple classifiers that enables it to adopt small networks for easy samples while switch to large models for hard ones. Another approach is to ensemble multiple models, and selectively execute a subset of them in the cascading \cite{bolukbasi2017adaptive, wang2021not} or mixing \cite{shazeer2017outrageously, ruiz2019adaptative} paradigm. Some other works propose to dynamically skip unnecessary layers \cite{veit2018convolutional, wang2018skipnet, wu2018blockdrop} or channels \cite{lin2017runtime}.

\subsection{Spatial redundancy}
Recent research has revealed that considerable spatial redundancy occurs when inferring deep networks \cite{figurnov2017spatially, xie2020spatially, Wang_2021_ICCV, wang2021adafocus, 9609974}. Several approaches have been proposed to reduce the redundant computation in the spatial dimension. The OctConv \cite{chen2019drop} reduces the spatial resolution by using low-frequency features. The Spatially Adaptive Computation Time (SACT) \cite{figurnov2017spatially} dynamically adjusts the number of executed layers for different image regions. The methods proposed in \cite{hua2019channel} and \cite{xie2020spatially} skip the computation on some less important regions of feature maps. These works mainly reduce the spatial redundancy by modifying convolutional layers, while we propose to process the image in a sequential manner. Our method is general as it does not require altering the network architecture.

\subsection{Visual attention models} 
Our GFNet is related to the visual attention models, which are similar to the human perception in that human usually pay attention to parts of the environment to perform recognition. Many existing works integrate the attention mechanism into image processing systems, especially in language-related tasks. For example, in image captaining and visual question answering, models are trained to concentrate on the related regions of the image when generating the word sequence \cite{xu2015show, vinyals2015show, vedantam2015cider, karpathy2015deep, yang2016stacked}. 
For image recognition, the attention mechanism is typically exploited to extract information from some task-relevant regions \cite{larochelle2010learning, ba2014multiple, jaderberg2015spatial, chu2019spot}.

One similar work to our GFNet is the recurrent visual attention model proposed in \cite{mnih2014recurrent}. However, our method differs from it in two important aspects: 1) we adopt a flexible and general CNN-based framework that is compatible with a wide variety of CNNs to achieve state-of-the-art computational efficiency, instead of sticking to a pure RNN model; and 2) our network focuses on performing adaptive inference for higher efficiency, and the recurrent process can be terminated conditioned on each input. With these design innovations, the proposed GFNet has achieved state-of-the-art performance on ImageNet in terms of both the theoretical computational efficiency and actual inference speed. In addition, \cite{NIPS2018_7411} shares a similar spirit to us in selecting important features with reinforcement learning, but it is not based on CNNs nor image data.


\begin{figure}[!t]
    \begin{center}
    \centerline{\hskip 0.1in \includegraphics[width=\columnwidth]{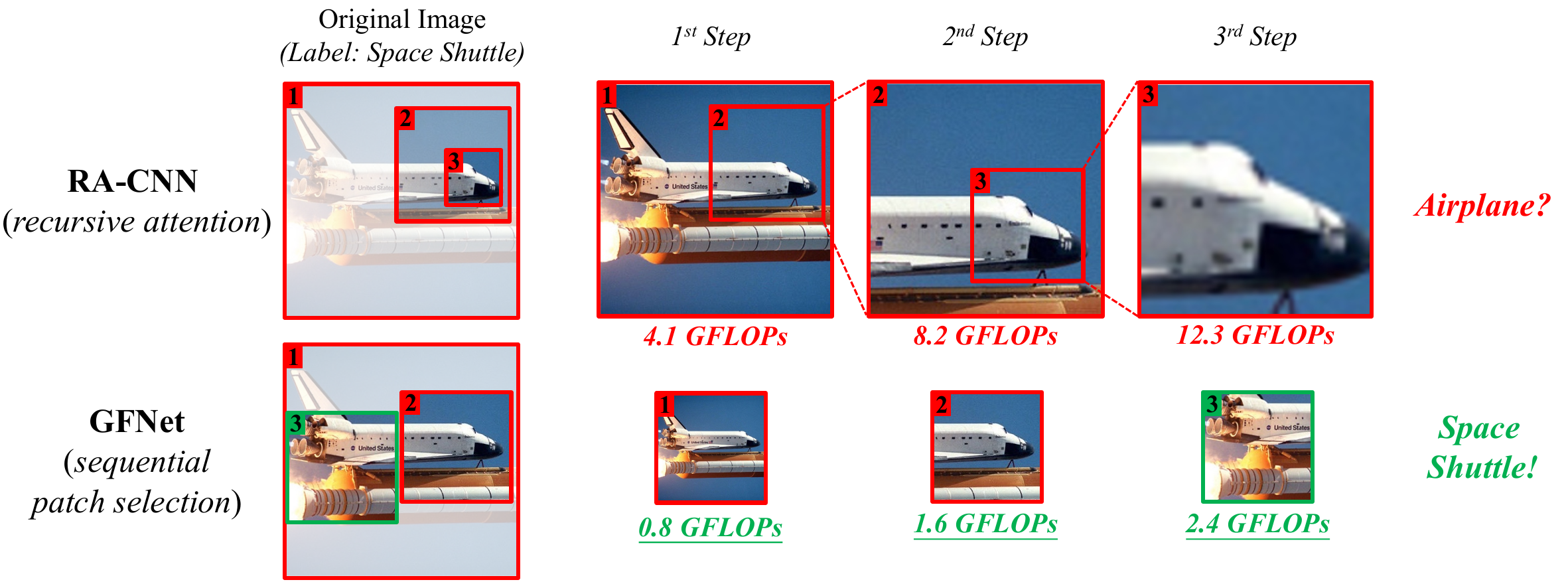}}
    \vspace{-2.5ex}
    \caption{The comparison of GFNet and RA-CNN \cite{fu2017look}. As a representative example, the computational cost (\emph{i.e.}, FLOPs) is computed based on the standard setting on ImageNet \cite{5206848} with the ResNet-50 \cite{He_2016_CVPR} backbone network, where GFNet adopts 96x96 patches. GFNet is computationally more efficient, and can capture the class-discriminative regions in any shape or size flexibly. In contrast, RA-CNN \cite{fu2017look} is designed to attend to the discriminative patterns concentrated in a small region of the original images. This mechanism is tailored for the fine-grained recognition task, but tends to be sub-optimal in more general and complex scenarios (e.g., on the large-scale comprehensive visual datasets like ImageNet). 
    }
    \label{fig:GF_vs_RA}
    \end{center}
    \vspace{-3ex}
\end{figure}

\subsection{Auto-zooming-in Approaches}
In particular, some existing methods propose to automatically identify and attend to certain informative details of the input images \cite{fu2017look, zheng2017learning, recasens2018learning, najibi2019autofocus, zhu2019ta, chen2019looks, wang2020looking}. As representative examples, RA-CNN \cite{fu2017look}, TASN \cite{zheng2019looking} and NTS-Net \cite{yang2018learning} learn to distinguish between the fine-grained categories with subtle visual differences by zooming in on the discriminative local patterns (\emph{e.g.}, the afterbrain colors of birds). S3N \cite{ding2019selective} and MGE-CNN \cite{zhang2019learning} leverage the class activation maps (CAMs) \cite{zhou2016learning, selvaraju2017grad} to localize and highlight the visual evidence for fine-grained image recognition. This idea is also explored in the context of vehicle re-identification \cite{he2019part}, image translation \cite{ma2018gan} and zero-shot recognition \cite{li2018discriminative, zhu2019semantic, chen2019hybrid}.

In this direction, a similar work to GFNet may be RA-CNN \cite{fu2017look}. However, GFNet is fundamentally different from RA-CNN in terms of both our motivation and our technical contributions. First, the goal of GFNet is to improve the efficiency of deep networks. Our proposed ``\emph{Glance and Focus}'' design enables GFNet to localize and leverage the class-discriminative image regions efficiently with small inputs (\emph{e.g.}, the 96x96 down-sampled images or the local patches, utilizing 18\% of the FLOPs for processing the original 224x224 image). On the contrary, RA-CNN \cite{fu2017look} is computationally intensive since it follows a different ``\emph{global-to-local}'' principle from us, \emph{i.e.}, all samples need to be first processed in high-resolution (\emph{e.g.}, 224x224), while the most informative region within the inputs will be cropped, amplified (\emph{e.g.}, to 224x224), and adopted as the new inputs recursively. Moreover, on top of our advantage in efficiency, we introduce a novel adaptive inference algorithm, which further significantly reduces the overall computational cost for inference. 

Second, GFNet is inherently able to capture the class-discriminative regions in any shape or size flexibly by selecting a sequence of small task-relevant patches. In contrast, RA-CNN \cite{fu2017look} is designed to recursively crop the informative sub-regions from the input image patch, narrow down the attention window progressively, and finally attend to the discriminative patterns concentrated in a small region of the original images. This mechanism is tailored for the fine-grained visual recognition task, but is sub-optimal in terms of the more general and complex scenarios we consider (\emph{e.g.}, on the large-scale comprehensive image/video benchmarks like ImageNet and Something-Something V1/V2). An illustration of this issue is presented in Figure \ref{fig:GF_vs_RA}.

Third, in our formulation, the training of the patch proposal network requires exploring all potential task-relevant regions across the whole images, which cannot be achieved by the training techniques proposed in \cite{fu2017look}. By contrast, we address this issue by developing a novel reinforcement learning based training algorithm. As we validate with extensive empirical results, our algorithm yields an effective and flexible patch selection policy.

\begin{figure*}[!t]
    \begin{center}
    \centerline{\includegraphics[width=1.89\columnwidth]{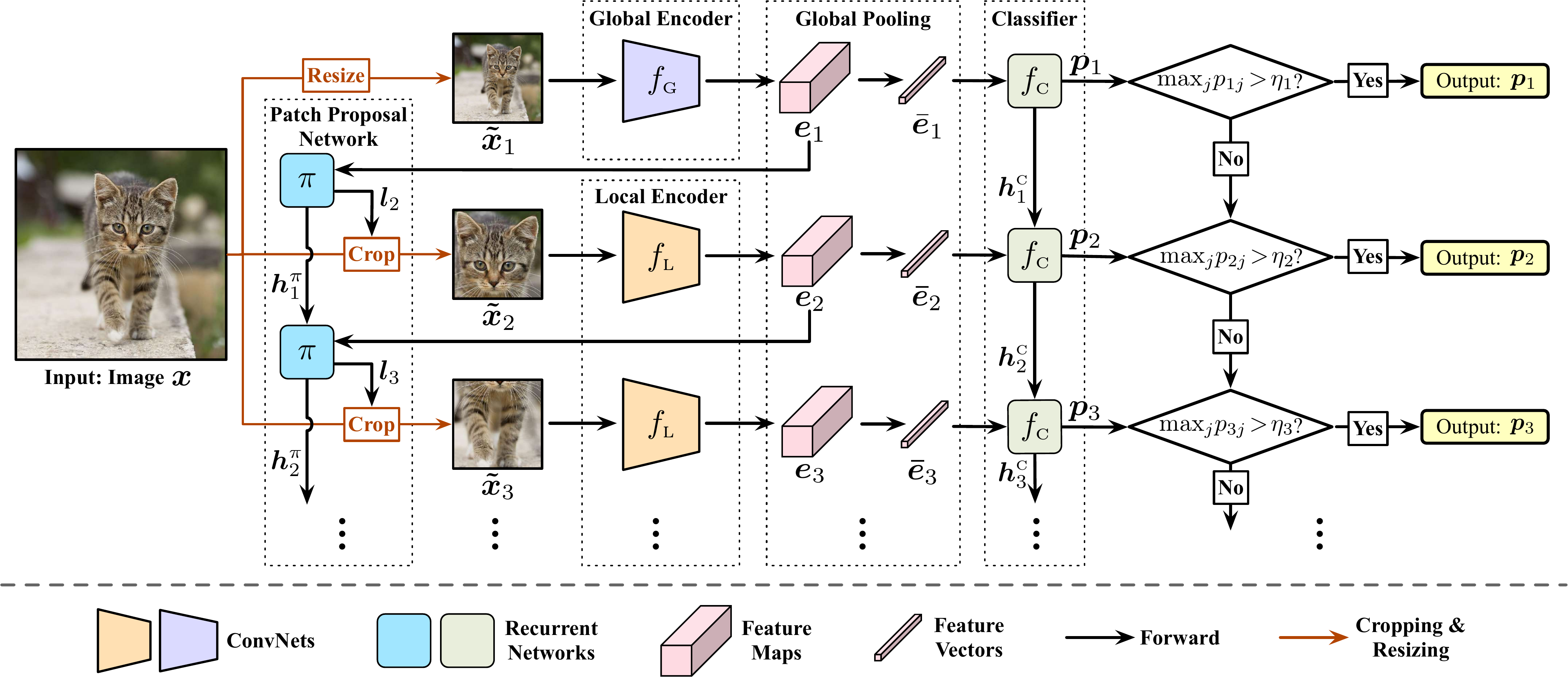}}
    \vspace{-1.5ex}
    \caption{
        An overview of GFNet. Given an input image $\bm{x}$, the model iteratively processes a sequence of patches $\{\tilde{\bm{x}}_1, \tilde{\bm{x}}_2, \dots\}$. The first input $\tilde{\bm{x}}_1$ is a low-resolution version of $\bm{x}$, and it is processed by the global encoder $f_{\textsc{g}}$ (\emph{Glance Step}). The following ones $\{\tilde{\bm{x}}_2, \dots\}$ are high-resolution patches cropped from $\bm{x}$, which are fed into the local encoder $f_{\textsc{l}}$ (\emph{Focus Stage}). At each step, GFNet produces a prediction with a classifier $f_{\textsc{c}}$, as well as decides the location of the next image patch using a patch proposal network $\pi$. This sequential decision process is terminated once sufficient confidence is obtained.
    }
    \label{fig:overview}
    \end{center}
    \vspace{-2ex}
\end{figure*}

%% file: method.tex
\section{Glance and Focus Networks (GFNet)}
In this section, we introduce the details of our method. For the ease of understanding, here we consider the task of recognizing images. We will show in Section \ref{sec:experiments} that it is effective to adopt the same techniques for processing videos.

As aforementioned, deep networks like CNNs are capable of producing accurate image classification results with certain ``class-discriminative'' image regions, such as the face of a dog or the wings of a bird. Inspired by this observation, we propose a GFNet framework, aiming to improve the computational efficiency of the models by performing computation on the minimal image regions to obtain a reliable prediction. To be specific, GFNet allocates computation adaptively and progressively to different areas of an image according to their contributions to the task of interest, while during inference, this process will be terminated once the network is adequately confident. 

\subsection{Overview}
In this subsection, we start by giving an overview of the proposed GFNet (as shown in Figure \ref{fig:overview}). Details of its components will be presented in Section \ref{sec:arch}.

Given an image $\bm{x}$ with the size $H\!\times\!W$, our method processes it with a sequence of $H'\!\times\!W'$ smaller inputs $\{\tilde{\bm{x}}_1, \tilde{\bm{x}}_2, \dots\}$, where $H'\!<\!H, W'\!<\!W$. These inputs are image patches cropped from certain locations of the image (except for $\tilde{\bm{x}}_1$, which will be described later). The specific location of each patch is dynamically determined by the network using the information of all previous inputs.

Ideally, the contributions of the inputs to classification should be descending in the sequence, such that the computational resources are first spent on the most valuable regions. However, given an arbitrary image, we do not have any specific prior knowledge on which regions are more important when generating the first patch $\tilde{\bm{x}}_1$. Therefore, we simply resize the original image $\bm{x}$ to $H'\!\times\!W'$ as $\tilde{\bm{x}}_1$, which not only avoids the risk of wasting computation on less important regions caused by randomly localizing the initial region, but also provides necessary global information that is beneficial for determining the locations of the following patches.


\textbf{Inference.}
The inference procedure of GFNet is formulated as a dynamic decision process on top of the input sequence $\{\tilde{\bm{x}}_1, \tilde{\bm{x}}_2, \dots\}$. At $t^{\textnormal{th}}$ step, the backbone feature extractor ($f_G$ or $f_L$) receives the input $\tilde{\bm{x}}_t$, and the model produces a softmax prediction $\bm{p}_t$. Then the largest entry of $\bm{p}_t$, \emph{{i.e.}}, $\max_j p_{tj}$, (treated as confidence following earlier works \cite{huang2017multi,yang2020resolution}) is compared to a pre-defined threshold $\eta_t$. If $\max_j p_{tj} > \eta_t$, then the sequential process halts, and $\bm{p}_t$ will be output as the final prediction. Otherwise, the location of the next image patch $\tilde{\bm{x}}_{t+1}$ will be decided, and $\tilde{\bm{x}}_{t+1}$ will be cropped from the image as the input at the $(t+1)^{\textnormal{th}}$ step. Note that the prediction $\bm{p}_t$ and the location of $\tilde{\bm{x}}_{t+1}$ are obtained using two recurrent networks, such that they exploit the information of all previous inputs $\{\tilde{\bm{x}}_1, \tilde{\bm{x}}_2, \dots, \tilde{\bm{x}}_t\}$. The maximum length of the input sequence is restricted to $T$ by setting $\eta_T \!=\! 0$, while other confidence thresholds $\eta_t (1\!\leq\!t\!\leq\!T\!-\!1)$ are determined under the practical requirements for a given computational budget. Details on obtaining the thresholds are presented in Section \ref{sec:details}.


\textbf{Training.}
During training, we inactivate early-terminating by setting $\eta_t\!=\!1 (1\!\leq\!t\!\leq\!T-1)$, and enforce all prediction $\bm{p}_t$'s $(1 \leq t \leq T)$ to be correct with high confidence. For patch localization, we train the network to select the patches that maximize the increments of the softmax prediction on the ground truth labels between the adjacent two steps. In other words, we seek to find the most class-discriminative image patches that have not been seen by the network. This procedure exploits a policy gradient algorithm to address the non-differentiability.

\subsection{The GFNet Architecture}
\label{sec:arch}
The proposed GFNet consists of four components: a global encoder $f_{\textsc{g}}$, a local encoder $f_{\textsc{l}}$, a classifier $f_{\textsc{c}}$ and a patch proposal network $\pi$.


\textbf{Global encoder $f_{\textsc{g}}$ and local encoder $f_{\textsc{l}}$} are both backbone networks that we utilize to extract deep representations from the inputs. They share the same network architecture but with different parameters. The former is applied to the resized original image $\tilde{\bm{x}}_1$, while the later is applied to the selected image patches. We use two networks instead of one because we find that there is a discrepancy between the scale of the low-resolution inputs $\tilde{\bm{x}}_1$ and the high-resolution local patches, which leads to degraded performance with a single encoder (detailed results are given in Section \ref{subsec:ablation}).

\textbf{Classifier $f_{\textsc{c}}$} is a recurrent network that aggregates the information from all previous inputs and produces a prediction at each step. We assume that the $t^{\textnormal{th}}$ input $\tilde{\bm{x}}_t$ is fed into the encoder, obtaining the corresponding feature maps $\bm{e}_t$. We perform the global average pooling on $\bm{e}_t$ to get a feature vector $\bm{\bar{e}}_t$, and produce the prediction $\bm{p}_t$ by 
\begin{equation}
    \bm{p}_t = f_{\textsc{c}}(\bm{\bar{e}}_t, \bm{h}^{\textsc{c}}_{t-1}),
\end{equation}
where $\bm{h}^{\textsc{c}}_{t-1}$ is the hidden state of $f_{\textsc{c}}$, which is updated at the $(t-1)^{\textnormal{th}}$ step. Note that it is unnecessary to maintain the feature maps for the \emph{classifier} as classification usually does not rely on the spatial information they contain. The recurrent classifier $f_{\textsc{c}}$ and the aforementioned two encoders $f_{\textsc{g}}$, $f_{\textsc{l}}$ are trained simultaneously with the following classification loss:
\begin{shrinkeq}{0ex}
\begin{equation}
    \label{eq:loss_cls}
    \mathcal{L}_{\textnormal{cls}} = \frac{1}{|\mathcal{D}_{\textnormal{train}}|}\sum\nolimits_{(\bm{x}, y) \in \mathcal{D}_{\textnormal{train}}} \left[ \frac{1}{T}\sum\nolimits_{t=1}^{T} L_{\textnormal{CE}}(\bm{p}_t, y)\right].
\end{equation}
\end{shrinkeq}
Herein, $\mathcal{D}_{\textnormal{train}}$ is the training set, $y$ denotes the label corresponding to $\bm{x}$ and $T$ is the maximum length of the input sequence. We use the standard cross-entropy loss function $L_{\textnormal{CE}}(\cdot)$ during training.

\begin{wrapfigure}{r}{4cm}
    \vskip -0.26in
    \begin{minipage}[t]{\linewidth}
    \centering
    \includegraphics[width=\textwidth]{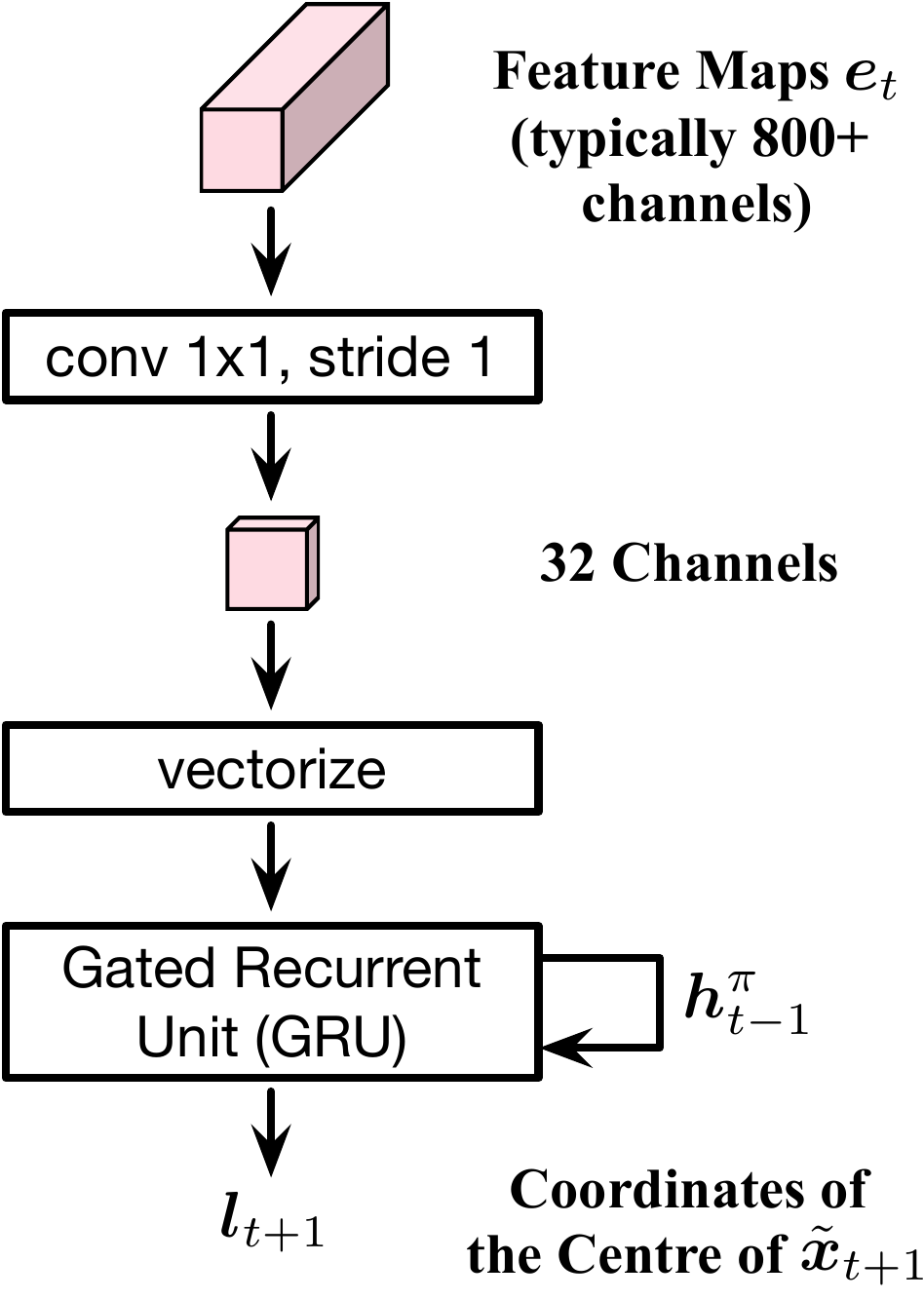}	
    \vskip -0.15in
    \caption{The architecture of the patch proposal network $\pi$. \label{fig:PPN}}  
    \end{minipage}
    \vskip -0.05in
\end{wrapfigure}

\textbf{Patch proposal network $\pi$} is another recurrent network that determines the location of each image patch. Given that the outputs of $\pi$ are used for the non-differentiable cropping operation, we model $\pi$ as an agent and train it using the policy gradient method. In specific, it receives the feature maps $\bm{e}_t$ of $\tilde{\bm{x}}_t$ at $t^{\textnormal{th}}$ step, and chooses a localization action $\bm{l}_{t+1}$ stochastically from a distribution parameterized by $\pi$: 
\begin{equation}
    \label{eq:action}
    \bm{l}_{t+1} \!\sim\! \pi(\bm{l}_{t+1}|\bm{e}_t, \bm{h}^{\pi}_{t-1}),
\end{equation}
where $\bm{l}_{t+1} \in [0, 1]^2$ is formulated as the normalized coordinates of the centre of the next patch $\tilde{\bm{x}}_{t+1}$. Here we use a Gaussian distribution during training,
whose mean is output by $\pi$ and standard deviation is pre-defined as a hyper-parameter. At test time, we simply adopt the mean value as $\bm{l}_{t+1}$ for a deterministic inference process. We denote the hidden state maintained within $\pi$ by $\bm{h}^{\pi}_{t-1}$, which aggregates the information of all past feature maps $\{\bm{e}_1, \dots, \bm{e}_{t-1}\}$. Note that we do not perform any pooling on $\bm{e}_t$, since the spatial information in the feature maps is essential for localizing the discriminative regions. On the other hand, we save the computational cost by reducing the number of feature channels using a 1$\times$1 convolution. Such a design abandons parts of the information that are valuable for classification but unnecessary for localization. The architecture of $\pi$ is shown in Figure \ref{fig:PPN}.

During training, after obtaining $\bm{l}_{t+1}$, we crop the $H'\!\times\!W'$ area centred at $\bm{l}_{t+1}$ from the original image $\bm{x}$ as the next input $\tilde{\bm{x}}_{t+1}$, and feed it into the network to produce the prediction $\bm{p}_{t+1}$. Then the patch proposal network $\pi$ receives a reward $r_{t+1}$ for the action $\bm{l}_{t+1}$, which is defined as the increments of the softmax prediction probability on the ground truth labels, \emph{{i.e.}}, 
\begin{equation}
    \label{eq:incre_reward}
    r_{t+1} = p_{(t+1)y} - p_{ty},
\end{equation}
where $y \in \{1, \dots, C\}$ is the label of $\bm{x}$ among $C$ classes. The goal of $\pi$ is to maximize the sum of the discounted rewards: 
\begin{shrinkeq}{0ex}
\begin{equation}
    \label{eq:reward}
    \max_{\pi}\ \  \mathbb{E}\left[ \sum\nolimits_{t=2}^{T} \gamma^{t-2} r_t \right] ,
\end{equation}
\end{shrinkeq}
where $\gamma \in (0,1)$ is a pre-defined discount factor. Intuitively, through Eq. (\ref{eq:reward}), we enforce $\pi$ to select the patches that enable the network to produce correct predictions in high confidence with as fewer patches as possible. In essence, we train $\pi$ to predict the location of the most beneficial region for the image classification at each step. Note that this procedure considers the previous inputs as well, since we compute the ``increments'' of the prediction probability.

\subsection{Training Strategy}
To ensure GFNet is trained properly, we propose a 3-stage training scheme, where the first two stages are indispensable, and the third stage is designed to further improve the performance.

\textbf{Stage \uppercase\expandafter{\romannumeral1}: }
At first, we do not integrate the patch proposal network $\pi$ into GFNet. Instead, we randomly crop the patch at each step with a uniform distribution over the entire input image, and train $f_{\textsc{g}}$, $f_{\textsc{l}}$ and $f_{\textsc{c}}$ to minimize the classification loss $\mathcal{L}_{\textnormal{cls}}$ (Eq. (\ref{eq:loss_cls})). In this stage, the network is trained to adapt to arbitrary input sequences.

\textbf{Stage \uppercase\expandafter{\romannumeral2}: }
We fix the two encoders and the classifier obtained from Stage \uppercase\expandafter{\romannumeral1}, and evoke a randomly initialized patch proposal network $\pi$ to decide the locations of image patches. Then we train $\pi$ using a policy gradient algorithm to maximize the total reward (Eq. (\ref{eq:reward})). 

\textbf{Stage \uppercase\expandafter{\romannumeral3}: }
Finally, we fine-tune the two encoders and the classifier with the fixed $\pi$ from Stage \uppercase\expandafter{\romannumeral2} to improve the accuracy of GFNet with the learned patch selection policy.

\subsection{Implementation Details}
\label{sec:details}
\textbf{Initialization of $f_{\textsc{g}}$ and $f_{\textsc{l}}$.}
We initialize the local encoder $f_{\textsc{l}}$ using the ImageNet pre-trained models. Since the global encoder $f_{\textsc{g}}$ processes the resized image with lower resolution, we first fine-tune the ImageNet pre-trained models with all training samples resized to $H'\!\times\! W'$, and then initialize $f_{\textsc{g}}$ with the fine-tuned parameters. 


\textbf{Recurrent networks.}
We adopt the gated recurrent unit (GRU) \cite{cho2014learning} in the classifier $f_{\textsc{c}}$ and the patch proposal network $\pi$. For MobileNets-V3 and EfficientNets, we use a cascade of fully-connected classifiers for efficient inference. The details are deferred to Appendix A.1.

\textbf{Regularizing Networks.}
In our implementation, we add a regularization term to Eq. (\ref{eq:loss_cls}), aiming to keep the capability of the two encoders to learn linearly separable representations, namely
\begin{shrinkeq}{0ex}
\begin{equation}
    \label{eq:loss_cls_impl}
    \begin{split}
        \mathcal{L}_{\textnormal{cls}}' = & \frac{1}{|\mathcal{D}_{\textnormal{train}}|}\sum\nolimits_{(\bm{x}, y) \in \mathcal{D}_{\textnormal{train}}} \left\{ \frac{1}{T}\sum\nolimits_{t=1}^{T} \left[
        L_{\textnormal{CE}}(\bm{p}_t, y) \right.\right. \\ 
        & \left.\left. + \lambda L_{\textnormal{CE}}(\textnormal{Softmax}(\textnormal{FC}_t(\bm{\bar{e}}_t)), y)
    \right] \right\},
    \end{split}
\end{equation}
\end{shrinkeq}
where $\lambda > 0$ is a pre-defined coefficient. Herein, we define a fully-connected layer $\textnormal{FC}_t(\cdot)$ for each step, and compute the softmax cross-entropy loss over the feature vector $\bm{\bar{e}}_t$ with $\textnormal{FC}_t$. Note that, when minimizing Eq. (\ref{eq:loss_cls}), the two encoders are not directly supervised since all gradients flow through the classifier $f_{\textsc{c}}$, while Eq. (\ref{eq:loss_cls_impl}) explicitly enforces a linearized deep feature space. 



\textbf{Confidence thresholds.}
One of the prominent advantages of GFNet is that both its computational cost and its inference latency can be tuned online according to the practical requirements via changing the confidence thresholds $\{\eta_1, \eta_2, \ldots\}$. To solve their values, we consider a \textit{budgeted batch classification} \cite{huang2017multi} scenario, where the model needs to classify a set of samples $\mathcal{D}_{\textnormal{test}}$ within a given computational budget $B\!>\!0$. Let $\textnormal{Accuracy}(\mathcal{D}_{\textnormal{val}}, \{\eta_1, \eta_2, \ldots\})$ and $\textnormal{FLOPs}(\mathcal{D}_{\textnormal{val}}, \{\eta_1, \eta_2, \ldots\})$ denote the accuracy and computational cost of GFNet on the validation set $\mathcal{D}_{\textnormal{val}}$ with $\{\eta_1, \eta_2, \ldots\}$. Then the thresholds can be obtained by solving the following optimization problem:
\begin{equation}
    \begin{split}
        \label{problem:threshold}
        \mathop{\textnormal{maximize}}_{\eta_1, \eta_2, \ldots}\ \ &\textnormal{Accuracy}(\mathcal{D}_{\textnormal{val}}, \{\eta_1, \eta_2, \ldots\}), \\
    s.t.\ \ &\textnormal{FLOPs}(\mathcal{D}_{\textnormal{val}}, \{\eta_1, \eta_2, \ldots\})\leq B.
    \end{split}
\end{equation}
Unless otherwise specified, problem (\ref{problem:threshold}) will be solved with the following procedure. We assume that the probability of obtaining the final prediction at $t^{\textnormal{th}}$ step is $q_t$, and the corresponding computational cost or latency is $C_t$. Then the average cost for each sample can be computed as $\sum_t q_t C_t$, leading to the constraint $|\mathcal{D}_{\textnormal{val}}|\sum_t\!q_t C_t\!\leq\!B$. We can solve this constraint for a proper $q_t$ and determine the threshold $\eta_t$ on the validation set. In our implementation, following \cite{huang2017multi}, we let $q_t\!=\!z (1-q)^{t-1}q$, where $z$ is a normalizing constant to ensure $\sum_t\!q_t\!=\!1$, and $0\!<\!q\!<\!1$ is the variable to be solved.




\textbf{Policy gradient algorithm.}
We implement the off-the-shelf proximal policy optimization (PPO) algorithm proposed by \cite{schulman2017proximal} to train the patch proposal network $\pi$. The details are introduced in Appendix A.2.

\begin{figure}[t]
    \begin{minipage}[t]{\linewidth}
    \centering
    \includegraphics[width=0.9\textwidth]{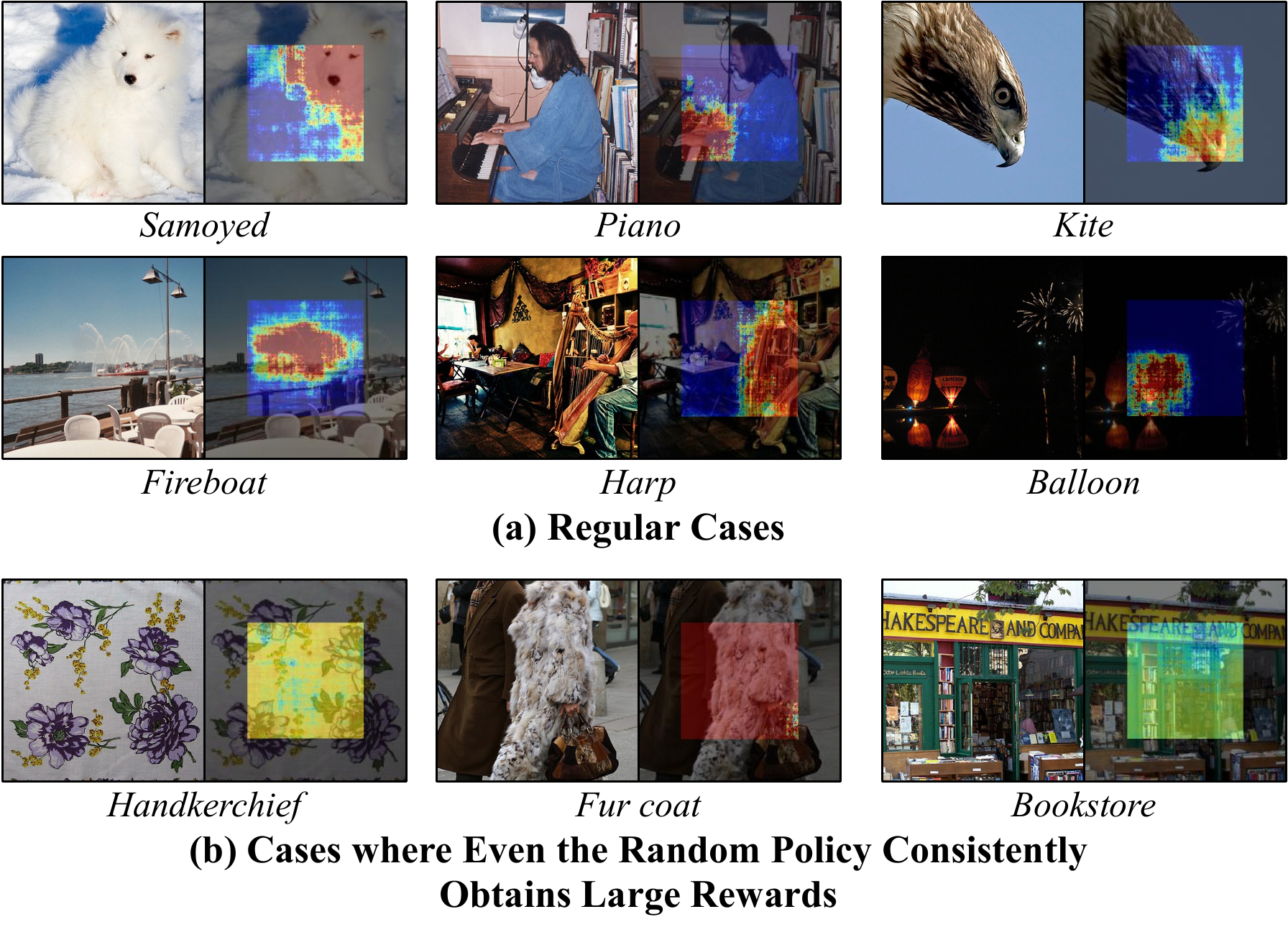}	
    \vskip -0.15in
    \caption{The value of rewards when the centre of the $2^{\textnormal{rd}}$ patch is located at each location in a GF-ResNet-50 ($T\!\!=\!\!5$, $H'\!\!=\!\!W'\!\!=\!96$), indicated by heat maps. The original images are shown on the left. We assume that the samples shown in (b) may confuse the learning of the patch selection policy using the reward proposed in Eq. (\ref{eq:incre_reward}), where all the actions will be encouraged.
    \label{fig:contrastive_reward}}  
    \end{minipage}
    \vskip -0.1in
\end{figure}

\section{Improved Techniques for GFNet}
\label{sec:improved_technique}
In this section, we propose a contrastive reward function and a resizable patch mechanism to further boost the computational efficiency of GFNet. The former improves the performance of the learned patch selection policy. The latter enables a single GFNet to achieve high accuracy when its computational cost is online tuned among a large range of budgets. The effectiveness of these two improved techniques is validated in Sections \ref{sec:exp_contrastive_reward} and \ref{sec:exp_msgf}.

\subsection{Contrastive Reward}
\label{sec:contrastive_reward}
With the reward proposed in Eq. (\ref{eq:incre_reward}), we ideally hope the patch proposal network $\pi$ can find the patches which maximize the confidence increments. However, it is empirically observed that in some cases, even selecting patches randomly may consistently lead to large rewards. Examples are shown in Figure \ref{fig:contrastive_reward} (b). The color in heat maps denotes the value of confidence increments when the centre of the second patch ($\tilde{\bm{x}}_{2}$) is located at each corresponding location. 

We assume that these instances may confuse the training of $\pi$, since every action will be encouraged due to its large reward. To alleviate this problem, we propose a contrastive reward function, \emph{i.e.},
\begin{equation}
    \label{eq:contrastive_reward}
    \begin{split}
        r_{t+1} \!=\ p_{(t+1)y} \!- {\mathbb{E}}_{\tilde{\bm{x}}_{t+1}\sim\textnormal{RandomCrop}(\bm{x})}\!\left[p_{(t+1)y}\right],
    \end{split}
\end{equation}
where ${\mathbb{E}}_{\tilde{\bm{x}}_{t+1}\sim\textnormal{RandomCrop}(\bm{x})}\!\left[p_{(t+1)y}\right]$ is the expected softmax probability on the ground truth label when the $(t+1)^{\textnormal{th}}$ input $\tilde{\bm{x}}_{t+1}$ is randomly cropped from the image. With Eq. (\ref{eq:contrastive_reward}), an action taken by $\pi$ will be compared with the average effect of a random policy to determine whether it should be encouraged. As a consequence, only the case where strategically selecting patches leads to a significant change of network prediction may produce the varying rewards. If all possible actions tend to be equally important, and hence learning $\pi$ makes less sense, the value of rewards will be around zero, avoiding confusing the training process. 

\begin{figure}[t]
    \begin{minipage}[t]{\linewidth}
    \centering
    \includegraphics[width=0.95\textwidth]{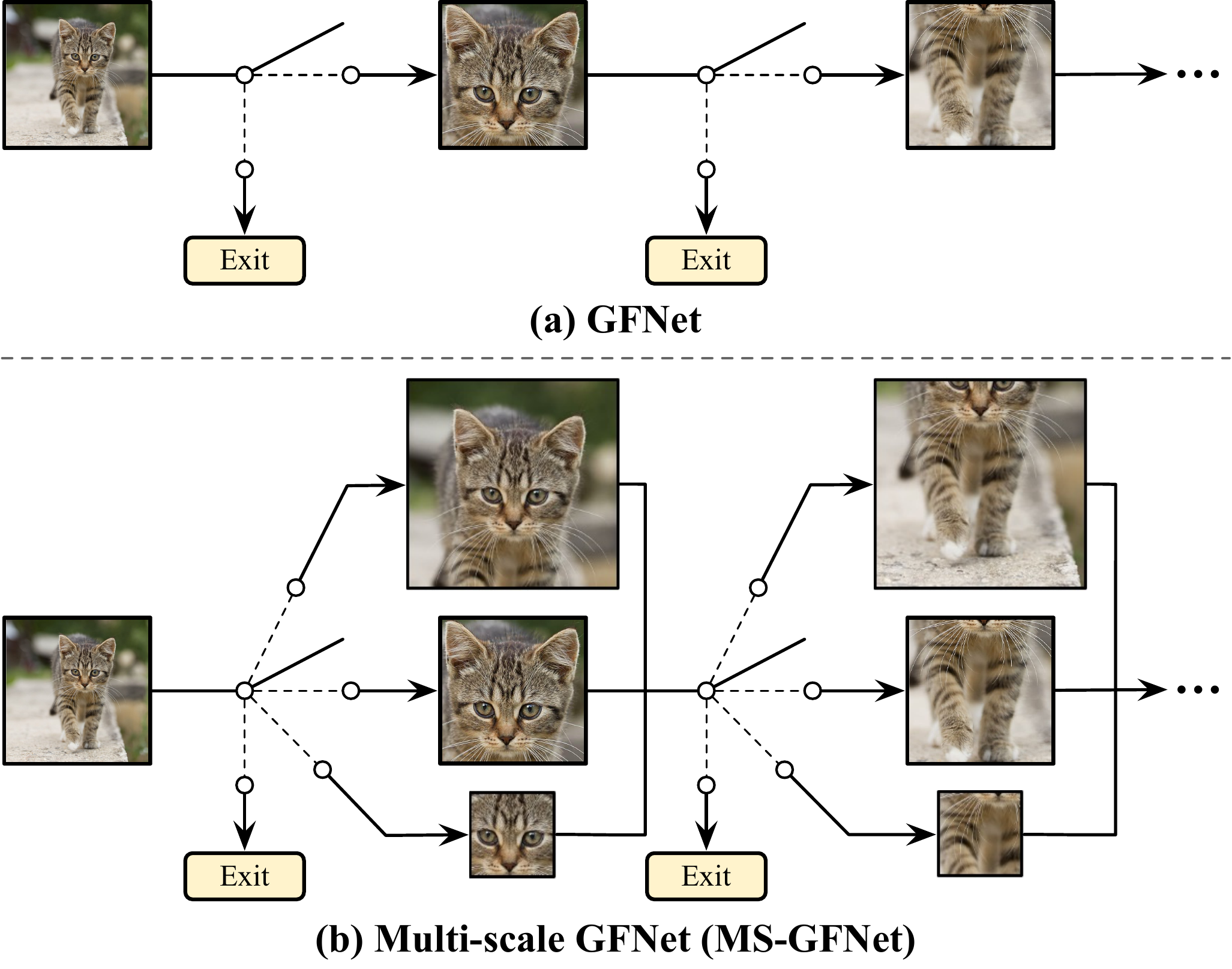}	
    \vskip -0.15in
    \caption{Comparisons of the vanilla GFNet and Multi-scale GFNet (MS-GFNet). The later allows the model to dynamically adjust the size of patches, such that high accuracy can be achieved with both small and relatively sufficient computational budgets. Note that the computational cost of both GFNet and MS-GFNet can be adjusted online without additional training.
    \label{fig:ms_gfnet}}  
    \end{minipage}
    \vskip -0.1in
\end{figure}

\subsection{Multi-scale GFNet}
\label{sec:MS-GFNet}
As aforementioned, the vanilla GFNet fixes the size of patches (\emph{i.e.}, $H'\!\times\!W'$) as a pre-defined hyper-parameter. However, this constraint limits the flexibility of a single GFNet to achieve high computational efficiency when its computational cost is online tuned within a large range. Given a small computational budget, adopting a smaller patch size usually results in higher accuracy than using larger patches. However, once the available computational resources are relatively sufficient, larger patches typically outperform smaller ones. The empirical results on this phenomenon can be found in Table \ref{tab:MS-GFNet}. 

Inspired by this issue, we propose a multi-scale GFNet (MS-GFNet), as shown in Figure \ref{fig:ms_gfnet}. At test time, after processing each image patch, MS-GFNet determines whether to output the prediction as the vanilla GFNet does. Once the inference proceeds, the size of the next patch will be selected among several candidates. Here we integrate these two decision steps into a single confidence-based criterion, which we empirically find effective. Formally, at the $t^{\textnormal{th}}$ step, we assume that the $k^{\textnormal{th}}$ candidate of patch size corresponds to a threshold $\eta_t^k$. Given the prediction $\bm{p}_t$, the inference process of the samples with $\max_j p_{tj} > \max_k \eta_t^k$ will be terminated, while the samples with 
\begin{equation}
    \max_j p_{tj} \leq \eta_t^k \ \ \ \textnormal{and}\ \ \  \max_j p_{tj} > \max_l \{\eta_t^l | \eta_t^l \leq \eta_t^k\}
\end{equation}
will use $k^{\textnormal{th}}$ patch size at $(t+1)^{\textnormal{th}}$ step, and repeat this procedure with $\{\eta_{t+1}^1, \eta_{t+1}^2, \ldots\}$. The values of all the thresholds can be treated as hyper-parameters and be determined on the validation set following problem (\ref{problem:threshold}), where we adopt the genetic algorithm \cite{whitley1994genetic}. As a result, both the input sequence lengths and the patch sizes can be dynamically adjusted conditioned on the computational budgets. MS-GFNet by design can switch among using smaller and larger patches online (without additional training) to achieve relatively better performance with either limited or sufficient computational resources. During training, we simply randomly select patch sizes in all the three stages. 

The architecture of the two encoders ($f_{\textsc{g}}$ and $f_{\textsc{l}}$) and the classifier ($f_{\textsc{c}}$) remains unchanged in MS-GFNet. For the former, most of deep networks can naturally process the inputs with varying sizes, while for the later, the dimension of the input feature vectors ($\bm{\bar{e}}_t$) does not change due to the global pooling. It is worth noting that the patch proposal network $\pi$ receives the feature maps ($\bm{e}_t$) in multiple sizes, which is intractable using the original network. To this end, we upsample all the feature maps to the maximum possible size after the 1x1 convolution in $\pi$. In addition, to provide the information on the sizes of the last and the next patches, we encode them in one-hot and concatenate them with the features before they are fed into the GRU.

%% file: experiments.tex
\section{Experiments}
\label{sec:experiments}
In this section, we empirically evaluate the effectiveness of the proposed GFNet on both image-based and video-based recognition tasks. Code and pre-trained models are available at \url{https://github.com/blackfeather-wang/GFNet-Pytorch}.

\subsection{Setups}

\subsubsection{Datasets}

\textbf{Image classification.} 
(1) ImageNet is a 1,000-class dataset from ILSVRC2012 \cite{5206848}, with 1.2 million images for training and 50,000 images for validation. We adopt the same data augmentation and pre-processing configurations as \cite{2019arXiv160806993H, He_2016_CVPR, wang2019implicit}. Unless otherwise specified, the resolution of images is set to 224x224.
(2) The Swedish traffic signs dataset \cite{larsson2011using, katharopoulos2019processing} consists of 747 training images and 684 test images, with a resolution of 960x1280. The samples are annotated according to the types of speed limit signs they contained or no speed limit. The same data augmentation techniques as \cite{katharopoulos2019processing} are adopted. 
For both datasets, we estimate the confidence thresholds of GFNet on the training set, since we find that it achieves nearly the same performance as cross-validation.


\begin{figure*}[!tbp]
    \centering
    \subfigure[MobileNet-V3]{\hspace{-0.1in}
    \begin{minipage}[t]{0.35\linewidth}
    \centering
    \includegraphics[width=\linewidth]{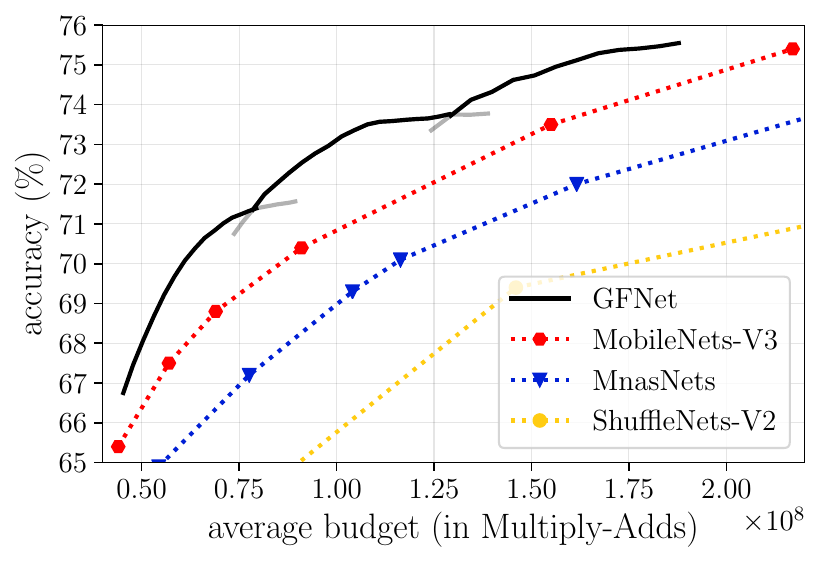}
    \end{minipage}%
    }\hspace{-0.1in}
    \subfigure[RegNet-Y]{
    \begin{minipage}[t]{0.33\linewidth}
    \centering
    \includegraphics[width=\linewidth]{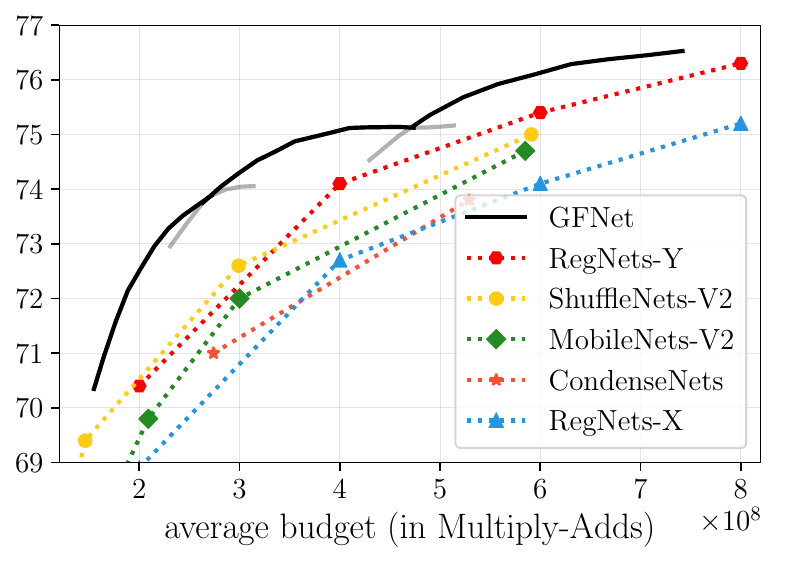}
    \end{minipage}%
    }\hspace{-0.1in}
    \subfigure[EfficientNet]{
    \begin{minipage}[t]{0.33\linewidth}
    \centering
    \includegraphics[width=\linewidth]{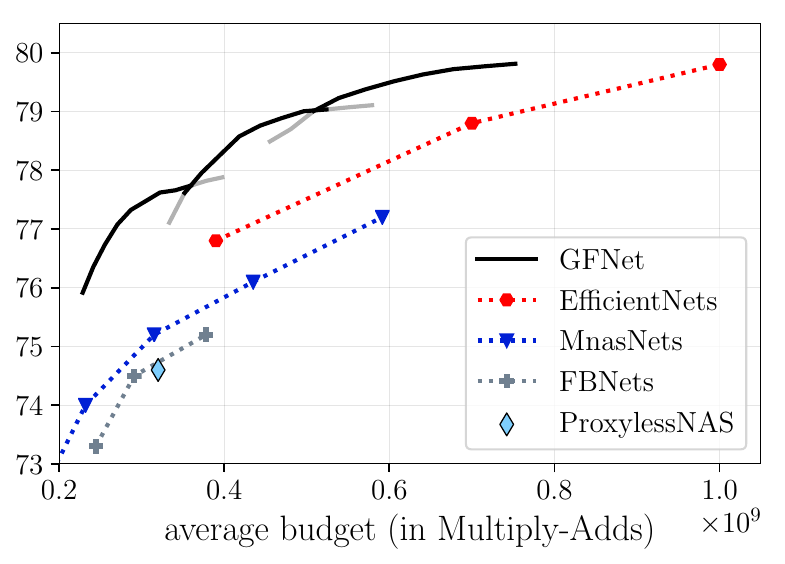}
    \end{minipage}%
    }%

    \vspace{-1ex} 
    \subfigure[ResNet]{\hspace{-0.065in}
    \begin{minipage}[t]{0.353\linewidth}
    \centering
    \includegraphics[width=\linewidth]{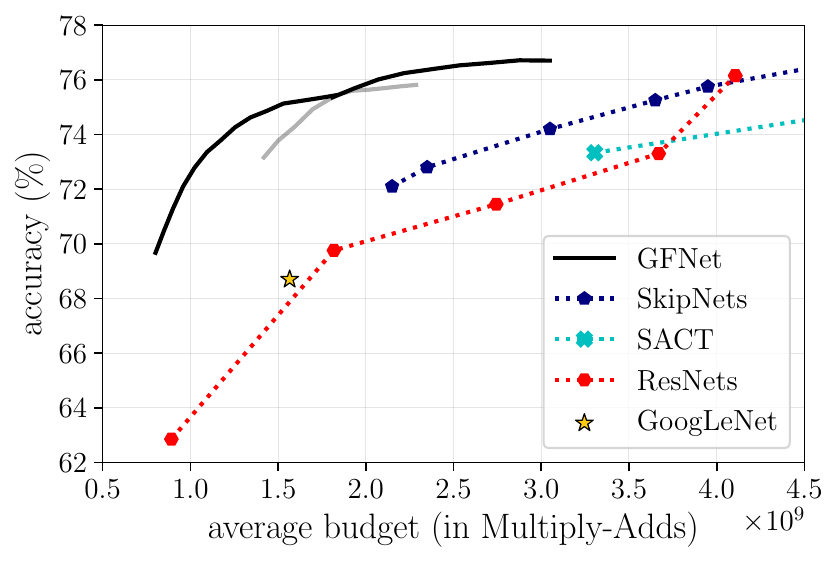}
    \end{minipage}
    }\hspace{-0.14in}
    \subfigure[DenseNet]{
    \begin{minipage}[t]{0.33\linewidth}
    \centering
    \includegraphics[width=\linewidth]{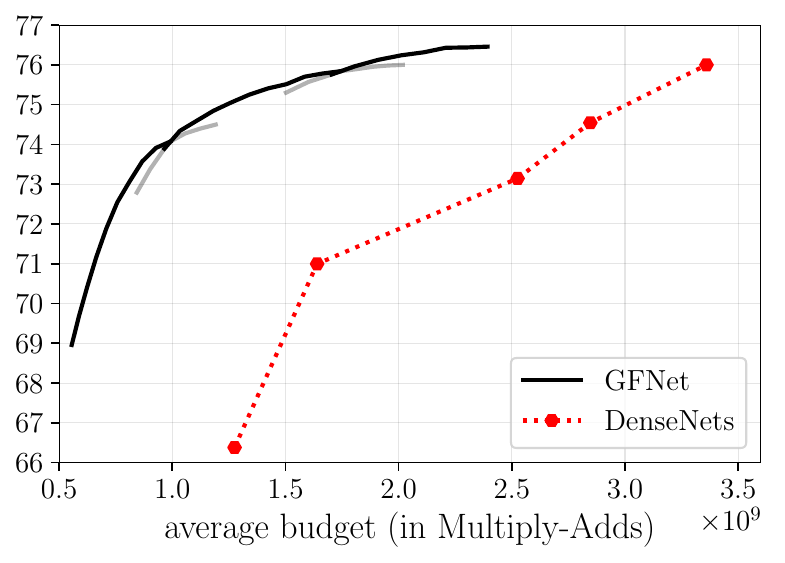}
    \end{minipage}
    }\hspace{-0.14in}
    \subfigure[DenseNet (anytime prediction)]{
    \begin{minipage}[t]{0.328\linewidth}
    \centering
    \includegraphics[width=\linewidth]{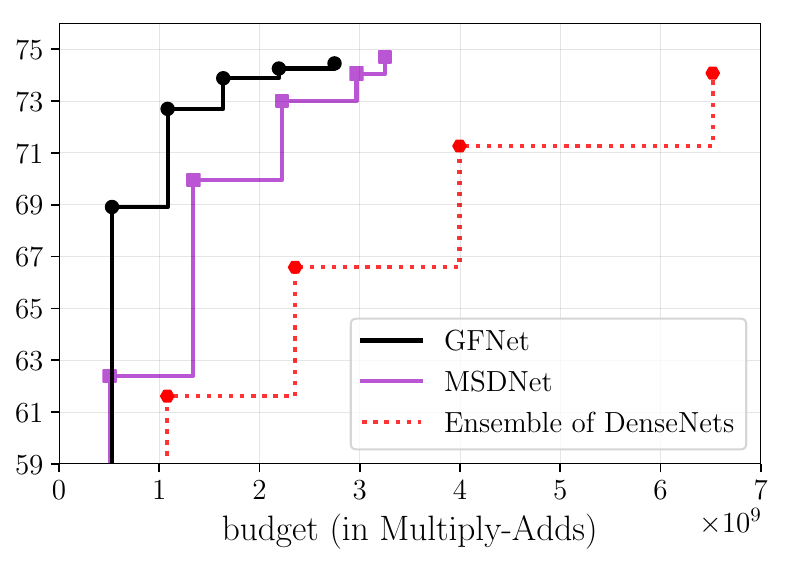}
    \end{minipage}%
    }
    \centering
    \vspace{-2.5ex}
    \caption{\label{fig:main_results}Top-1 accuracy v.s. Multiply-Adds on ImageNet. The proposed GFNet framework is implemented on top of state-of-the-art efficient networks. Figures (a-e) present the results of \textit{Budgeted batch classification}, while Figures (f) shows the \textit{Anytime prediction} results.}
    \vspace{-2ex}
\end{figure*}

\textbf{Video recognition.}
(1) Something-Something V1\&V2 \cite{goyal2017something} are two large-scale video recognition datasets, including 98k and 194k videos respectively. Both of them are annotated into 174 human action classes (e.g., pretending to pick something up).
(2) Jester \cite{materzynska2019jester} is a large-scale action recognition dataset consisting of 148,092 videos in 27 categories. 
We use the official training-validation split for all the three datasets, and uniformly sample 8/12/16 frames from each video. We use the same data augmentation and pre-processing configurations as \cite{lin2019tsm, Wang_2021_ICCV}. During inference, we resize all frames to 256x256 and centre-crop them to 224x224. The confidence thresholds are estimated on the training set.

\subsubsection{Two Inference Settings}
We consider two settings to evaluate our method: (1) \textit{budgeted batch classification} \cite{huang2017multi}, where the network needs to classify a set of test samples within a given computational budget; (2) \textit{anytime prediction} \cite{grubb2012speedboost, huang2017multi}, where the network can be forced to output a prediction at any given point in time. As discussed in \cite{huang2017multi}, these two settings are ubiquitous in many real-world applications. For (1), we estimate the confidence thresholds to perform the adaptive inference as introduced in Section \ref{sec:details}, while for (2), we assume the length of the input sequence is the same for all test samples.

\subsubsection{Backbone Networks}
For image classification, GFNet is implemented on the basis of several state-of-the-art CNNs, including MobileNet-V3 \cite{howard2019searching}, RegNet-Y \cite{radosavovic2020designing}, EfficientNet \cite{DBLP:conf/icml/TanL19}, ResNet \cite{He_2016_CVPR} and DenseNet \cite{2019arXiv160806993H}. These networks serve as the two deep encoders in our methods. In \textit{Budgeted batch classification}, for each sort of CNNs, we fix the maximum length of the input sequence $T$, and change the model size (width, depth or both) or the patch size ($H'$, $W'$) to obtain networks that cover different ranges of computational budgets. Note that we always let $H'\!=\!W'$. We also compare GFNet with a number of highly competitive baselines, \emph{i.e.}, MnasNets \cite{tan2019mnasnet}, ShuffleNets-V2 \cite{ma2018shufflenet}, MobileNets-V2 \cite{sandler2018mobilenetv2}, CondenseNets \cite{huang2018condensenet}, FBNets \cite{wu2019fbnet}, ProxylessNAS \cite{cai2018proxylessnas},  SkipNet \cite{wang2018skipnet}, SACT \cite{figurnov2017spatially}, GoogLeNet \cite{szegedy2015going} and MSDNet \cite{huang2017multi}.

For video recognition, we implement GFNet on top of a ResNet-50 with the temporal shift module (TSM) \cite{lin2019tsm}. With the aim of introducing minimum modifications, the resizing and cropping operations in GFNet are performed only in the spatial dimension in the same way as we do on images. 
Following \cite{lin2019tsm}, the whole video with all frames is always fed into the model. We fix $T$, and change $H'$, $W'$ (we always let $H'\!=\!W'$) and the number of frames sampled from each video among $\{8, 12, 16\}$. The performance of several efficient video recognition models are reported as baselines, including TRN \cite{zhou2018temporal}, ECO \cite{zolfaghari2018eco} and AdaFuse \cite{meng2021adafuse}.

More details on both network configurations and training hyper-parameters are presented in Appendices A.3 and A.4.


\subsection{Image Classification on ImageNet}

\subsubsection{Theoretical Computational Efficiency}
\label{sec:main_results}
\textbf{Budgeted batch classification}
results are shown in Figure \ref{fig:main_results} (a-e). We first plot the performance of each GFNet in a gray curve, and then plot the best validation accuracy with each budget as a black curve. It can be observed that GFNet significantly improves the performance of even the state-of-the-art efficient models with the same amount of computation. For example, with an average budget of $7\times10^7$ Multiply-Adds, the GFNet based on MobileNet-V3 achieves a Top-1 validation accuracy of $\sim71\%$, which outperforms the vanilla MobileNet-V3 by $\sim2\%$. With EfficientNets, GFNet generally has $\sim1.4\times$ less computation compared with baselines when achieving the same performance. With ResNets and DenseNets, GFNet reduces the number of required Multiply-Adds for the given test accuracy by approximately $2-3\times$ times. Moreover, the computational cost of our method can be tuned precisely to achieve the best possible performance with a given budget.

\textbf{Anytime prediction.}
We compare our method with another adaptive inference network, MSDNet \cite{huang2017multi}, under the \textit{Anytime prediction} setting in Figure \ref{fig:main_results} (f), where GFNet is based on a DenseNet-121. For fair comparison, here we hold out 50,000 training images, following \cite{huang2017multi} (we do not do so in Figure \ref{fig:main_results} (e), and the \textit{Budgeted batch classification} comparisons of GFNet and MSDNet are deferred to Appendix B.1). We also include the results of an ensemble of DenseNets with varying depth \cite{huang2017multi}. The plot shows that GFNet achieves $\!\sim\!4\!-\!10\%$ higher accuracy than MSDNet when the budget ranges from $5\times10^8$ to $2.2\times10^9$ Multiply-Adds.

\begin{figure*}
    \vspace{-0.025in}
    \begin{minipage}[t]{0.65\linewidth}
      \centering
    \hspace{-0.2in}
    \subfigure[MobileNet-V3]{\hspace{-0.1in}
      \begin{minipage}[t]{0.5\linewidth}
        \centering
        \includegraphics[width=\linewidth]{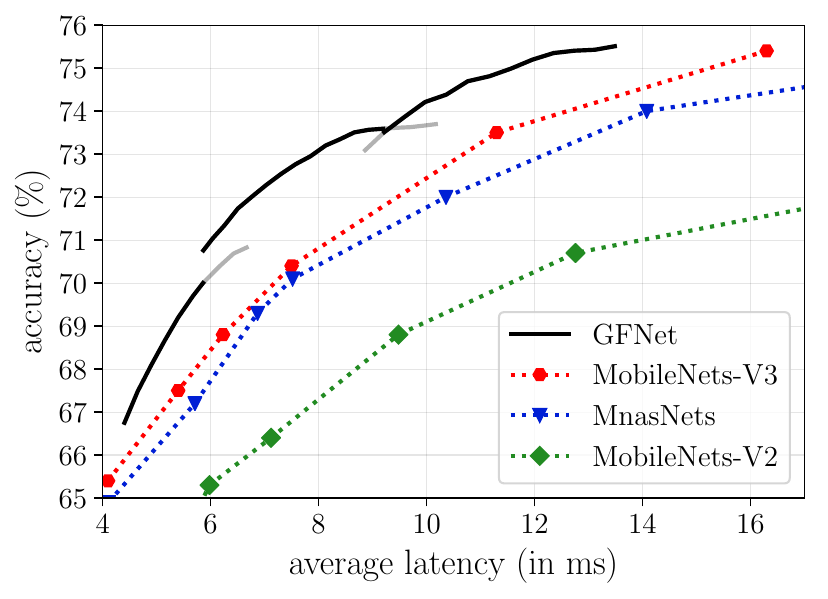}
      \end{minipage}}
      \subfigure[ResNet]{\hspace{-0.1in}
      \begin{minipage}[t]{0.4815\linewidth}
        \centering
        \includegraphics[width=\linewidth]{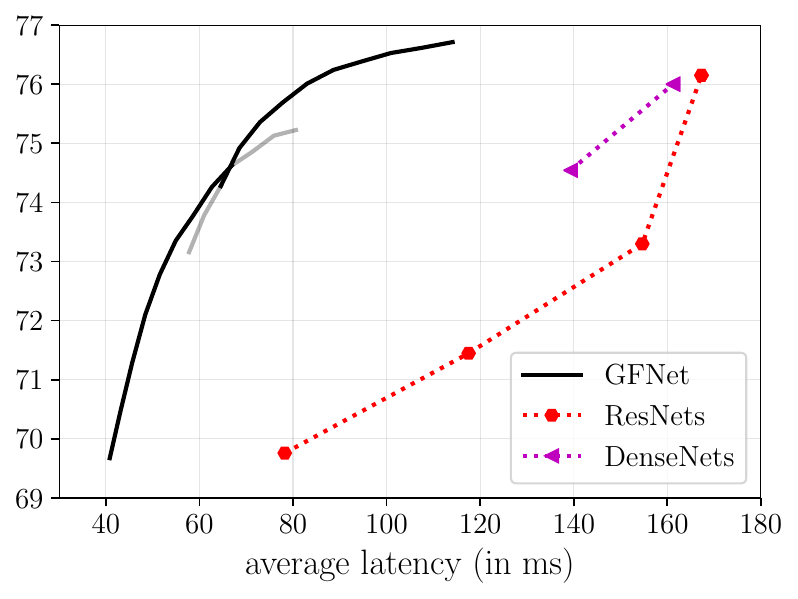}
      \end{minipage}}
      \vskip -0.175in
      \caption{\label{fig:speed} Top-1 accuracy v.s. inference latency (ms) on ImageNet. The inference speed is measured on an iPhone XS Max. We implement GFNet based on MobileNet-V3 and ResNet.}
    \end{minipage}
    \hspace{-0.03in}
    \begin{minipage}[t]{0.3465\linewidth}       
        \centering 
        \vskip -1.83in
        \begin{table}[H]
            \footnotesize
            \caption{Actual speed on GPU devices. We report the theoretical computational cost and the practical throughput when GFNet reaches the same Top-1 accuracy as its corresponding backbone CNN.}
            \vskip -0.2in
            \label{tab:gpu_speed}
            \setlength{\tabcolsep}{0.8mm}{
            \vspace{5pt}
            \renewcommand\arraystretch{1.25} 
            \resizebox{\columnwidth}{!}{
            \begin{tabular}{|c|c|c|c|}
                \hline

            \multirow{4}{*}{ResNet-50}
            &Average Budget & baseline & 4.11G  \\
            &(in Multiply-Adds) & GFNet & \textbf{2.16G}$_{\bm{{\textcolor{blue}{\downarrow1.90\times}}}}$    \\
             \cline{2-4}
             &Throughput & baseline & 976 img/s    \\
             &(on NVIDIA 2080Ti) & GFNet & \textbf{1812 img/s}$_{\bm{{\textcolor{blue}{\uparrow\!\!\!1.86\times}}}}$   \\
            \hline

            &Average Budget & baseline & 2.85G  \\
            DenseNet-&(in Multiply-Adds) & GFNet & \textbf{1.09G}$_{\bm{{\textcolor{blue}{\downarrow2.61\times}}}}$    \\
             \cline{2-4}
             -121&Throughput & baseline & 889 img/s    \\
             &(on NVIDIA 2080Ti) & GFNet & \textbf{1844 img/s}$_{\bm{{\textcolor{blue}{\uparrow\!\!\!2.07\times}}}}$   \\
             \hline

             &Average Budget & baseline & 3.36G  \\
             DenseNet-&(in Multiply-Adds) & GFNet & \textbf{1.83G}$_{\bm{{\textcolor{blue}{\downarrow1.84\times}}}}$    \\
              \cline{2-4}
              -169&Throughput & baseline & 722 img/s    \\
              &(on NVIDIA 2080Ti) & GFNet & \textbf{1086 img/s}$_{\bm{{\textcolor{blue}{\uparrow\!\!\!1.50\times}}}}$   \\
              \hline
            \end{tabular}}}
        \end{table}
    \end{minipage}
\end{figure*}

\begin{figure*}[t]
    \begin{center}
    \centerline{\includegraphics[width=2\columnwidth]{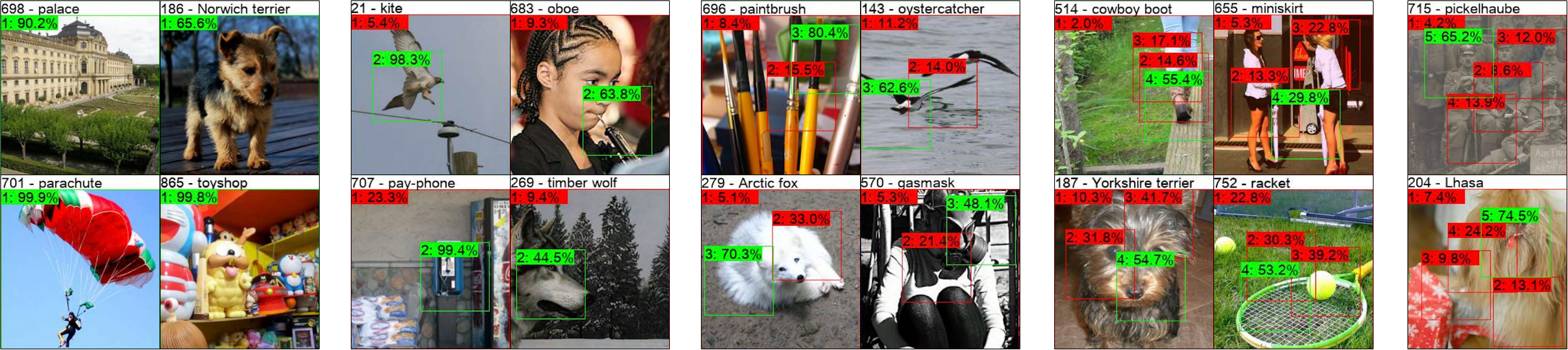}}
    \vspace{-1ex}
    \caption{Visualization results of the GF-ResNet-50 ($T\!\!=\!\!5$, $H'\!\!=\!\!W'\!\!=\!96$). The boxes indicate the patch locations, and the color denotes whether the prediction is correct at current step (green: correct; red: wrong). Note that $\tilde{\bm{x}}_1$ is the resized input image. The indices of the steps and the current confidence on the ground truth labels (shown at the top of images) are presented in the upper left corners of boxes. 
    }
    \label{fig:visualization}
    \end{center}
    \vspace{-2ex}
\end{figure*}

\subsubsection{Practical Inference Speed}

\textbf{Experiments on an iPhone.}
A suitable platform for the implementation of GFNet may be mobile applications, where the average inference latency and power consumption of each image are approximately linear in the amount of computation (Multiply-Adds) \cite{howard2019searching}, such that reducing computational costs helps for both improving user experience and preserving battery life. To this end, we investigate the practical inference speed of our method on an iPhone XS Max (with Apple A12 Bionic) using TFLite \cite{abadi2016tensorflow}. The single-thread mode with batch size 1 is used following \cite{howard2017mobilenets, sandler2018mobilenetv2, howard2019searching}. We first measure the time consumption of obtaining the prediction with each possible length of the input sequence, and then take the weighted average according to the number of validation samples using each length. The results are shown in Figure \ref{fig:speed}. One can observe that GFNet effectively accelerates the inference of both MobileNets-V3 and ResNets. For instance, our method reduces the required latency to achieve $75.4\%$ test accuracy (MobileNets-V3-Large) by $22\%$ (12.7ms v.s. 16.3ms). For ResNets, GFNet generally requires $2\!-\!3\times$ times lower latency to achieve the same performance as baselines.

\textbf{Experiments on GPUs.}
GFNet can also be leveraged to accelerate the inference of deep networks on GPU devices. Here we consider the batch inference setting, where a large number of test samples need to be processed using a GPU. In specific, the 50,000 images in the validation set of ImageNet are fed into the GFNet with a batch size of 128 on a single NVIDIA 2080Ti GPU. At each mini-batch, when the inference procedure reaches any exit (\emph{i.e.}, when producing $\bm{p}_1, \bm{p}_2, \ldots$), the samples that meet the early-termination criterion will be output, with the remaining images continuing to be processed (following the \emph{Focus Stage}). The throughput is computed by 50,000/$T$, where $T$ is the total wall-clock time of processing all the mini-batches. The results are shown in Table \ref{tab:gpu_speed}. With the same computational resources, GFNet improves the throughput of ResNet and DenseNet by $\sim\!1.5\!-\!2.1\times$ without sacrificing the accuracy.




\begin{table}[!t]
    \vskip -0.05in
    \caption{Comparisons of the contrastive reward with the original one. The results of three GFNets on ImageNet are reported. We present the Top-1 accuracy after the training stage \uppercase\expandafter{\romannumeral2} with fixed length of the input sequence (denoted by $t$). \label{tab:contrastive_vs_original}}
    \centering
    \vskip -0.125in
    \subtable[GF-ResNet-50 ($H'\!=\!W'\!=\!96$, $T\!=\!5$)]{
        \hspace{-0.125in}
        \setlength{\tabcolsep}{0.75mm}{
            \vspace{5pt}
            \renewcommand\arraystretch{1.25} 
            \begin{tabular}{|c|ccccc|}
            \hline
            Reward & $t=1$ & $t=2$ & $t=3$ & $t=4$ & $t=5$ \\
            \hline
            Increments &\ 69.62\%\ &73.23\%{\tiny{$\pm$0.13}}&74.24\%{\tiny{$\pm$0.11}}&74.84\%{\tiny{$\pm$0.18}}&75.25\%{\tiny{$\pm$0.08}} \\
            Contrastive &\ 69.62\%\ &\textbf{73.65\%}{\tiny{$\pm$0.17}}&\textbf{74.64\%}{\tiny{$\pm$0.06}}&\textbf{75.14\%}{\tiny{$\pm$0.09}}&\textbf{75.40\%}{\tiny{$\pm$0.09}} \\
            \hline
    \end{tabular}}
    }
     
    \qquad
     
    \subtable[GF-MobileNet-V3-Large ($H'\!=\!W'\!=\!128$, $T\!=\!3$)]{
        \setlength{\tabcolsep}{1mm}{
            \vspace{5pt}
            \renewcommand\arraystretch{1.25} 
            \begin{tabular}{|c|ccc|}
            \hline
            Reward & $t=1$ & $t=2$ & $t=3$  \\
            \hline
            Increments &\ 70.72\%&72.86\%{\tiny{$\pm$0.06}}&73.43\%{\tiny{$\pm$0.10}} \\
            Contrastive &\ 70.72\%&\textbf{73.30\%}{\tiny{$\pm$0.10}}&\textbf{73.72\%}{\tiny{$\pm$0.03}} \\
            \hline
    \end{tabular}}
    }
     
    \qquad
     
    \subtable[GF-EfficientNet-B2 ($H'\!=\!W'\!=\!128$, $T\!=\!4$)]{
        \setlength{\tabcolsep}{1mm}{
            \vspace{5pt}
            \renewcommand\arraystretch{1.25} 
            \begin{tabular}{|c|cccc|}
            \hline
            Reward & $t=1$ & $t=2$ & $t=3$ & $t=4$  \\
            \hline
            Increments &\ 75.39\%& 77.15\%{\tiny{$\pm$0.02}}& 77.73\%{\tiny{$\pm$0.06}} &77.90\%{\tiny{$\pm$0.03}} \\
            Contrastive &\ 75.39\%&\textbf{77.31\%}{\tiny{$\pm$0.04}}&\textbf{77.89\%}{\tiny{$\pm$0.05}} &\textbf{77.95\%}{\tiny{$\pm$0.03}} \\
            \hline
    \end{tabular}}
    }
    \vskip -0.2in
\end{table}

\subsubsection{Visualization}
We show the image patches found by a ResNet-50 based GFNet on some of test samples in Figure \ref{fig:visualization}. Samples are divided into different columns according to the number of inputs they require to obtain correct classification results. One can observe that GFNet classifies ``easy'' images containing large objects with prototypical features correctly at the \textit{Glance Step} with high confidence, while for relatively ``hard'' images which tend to be complex or non-typical, our network is capable of focusing on some class-discriminative regions to progressively improve the confidence.

\subsubsection{Effectiveness of the Contrastive Reward}
\label{sec:exp_contrastive_reward}

In Table \ref{tab:contrastive_vs_original}, we present the performance of the GFNets on top of ResNet-50, MobileNet-V3-Large and EfficientNet-B2 when the contrastive reward function proposed in Section \ref{sec:contrastive_reward} is used, where we estimate the expectation in Eq. (\ref{eq:contrastive_reward}) with a single time of Monte-Carlo sampling. The results of the original reward is referred to as ``Increments''. Different rewards are applied in training stage \uppercase\expandafter{\romannumeral2} (reinforcement learning) on the basis of the same stage \uppercase\expandafter{\romannumeral1} checkpoint. For a clear comparison, here we do not perform training stage \uppercase\expandafter{\romannumeral3}. One can observe that the proposed contrastive reward consistently improves the accuracy of all the three models, especially at the first step of the focus stage (\emph{i.e.}, $t=2$).

\begin{table*}[!t]
	\footnotesize
    \centering
    \caption{Top-1 accuracy of MS-GFNet (ResNet-50) on ImageNet under the \textit{Budgeted batch classification} setting. We fix the patch size of the \emph{Glance Step} (\emph{i.e.}, 96x96), and consider three candidate sizes in the \emph{Focus Stage} (\emph{i.e.}, 96x96, 128x128, 160x160). The performance of using fixed patch sizes in the \emph{Focus Stage} is also presented. The best result with each computational budget are \textbf{bold-faced}, which the second best result is \underline{underlined}.}
    \vskip -0.175in
    \label{tab:MS-GFNet}
    \setlength{\tabcolsep}{1.5mm}{
    \vspace{5pt}
    \renewcommand\arraystretch{1.3} 
    \begin{tabular}{|c|lllllllll|}
        \hline
    \multirow{2}{*}{Patch Size} & \multicolumn{9}{c|}{Average Budget (in Multiply-Adds)} \\
     &  \ 1.00G & 1.25G & 1.50G & 1.75G & 2.00G & 2.25G & 2.50G & 2.75G & 3.00G \\
    \hline
    96x96 (fixed) &\ \underline{73.28\%} & \underline{74.51\%} & 75.27\% & 75.59\% & 75.78\% & 75.90\% & -- & -- & -- \\
    128x128 (fixed) &  \ 72.65\% & 74.40\% & \underline{75.47\%} & \underline{76.07\%} & \underline{76.26\%} & 76.39\% & 76.40\% & 76.40\% & 76.40\% \\
    160x160 (fixed) &  \ 72.04\% & 73.58\% & 74.68\% & 75.48\% & 76.03\% & \underline{76.44\%} & \underline{76.72\%} & \underline{76.83\%} & \underline{76.94\%} \\
    \hline
    MS-GFNet &  \ \textbf{73.70\%}{\tiny{$\pm$0.04}} & \textbf{74.97\%}{\tiny{$\pm$0.06}} & \textbf{75.70\%}{\tiny{$\pm$0.08}} & \textbf{76.24\%}{\tiny{$\pm$0.07}} & \textbf{76.51\%}{\tiny{$\pm$0.07}} & \textbf{76.71\%}{\tiny{$\pm$0.08}} & \textbf{76.85\%}{\tiny{$\pm$0.06}} & \textbf{76.91\%}{\tiny{$\pm$0.05}} & \textbf{76.98\%}{\tiny{$\pm$0.09}} \\
    \hline
    \end{tabular}}
    \vskip -0.15in
\end{table*}

\begin{figure}[t]
    \vspace{-0.05in}
    \begin{minipage}[t]{\linewidth}
      \centering
    \hspace{-0.2in}
    \subfigure[ResNet]{\hspace{-0.1in}
      \begin{minipage}[t]{0.518\linewidth}
        \centering
        \includegraphics[width=\linewidth]{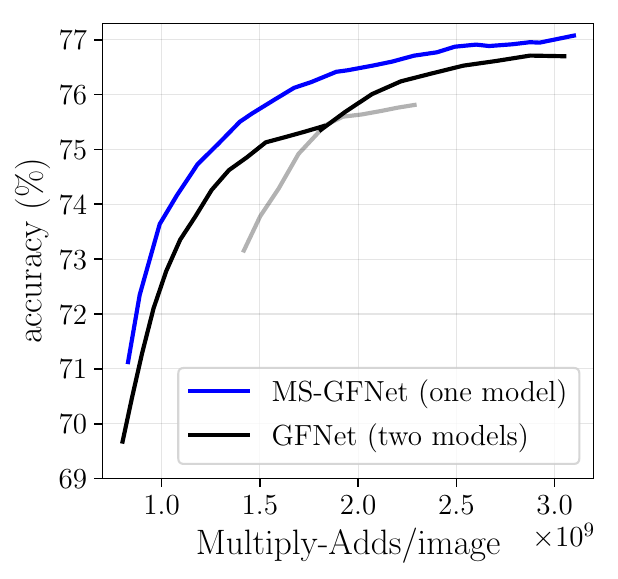}
      \end{minipage}}
      \subfigure[EfficientNet]{\hspace{-0.1in}
      \begin{minipage}[t]{0.4815\linewidth}
        \centering
        \includegraphics[width=\linewidth]{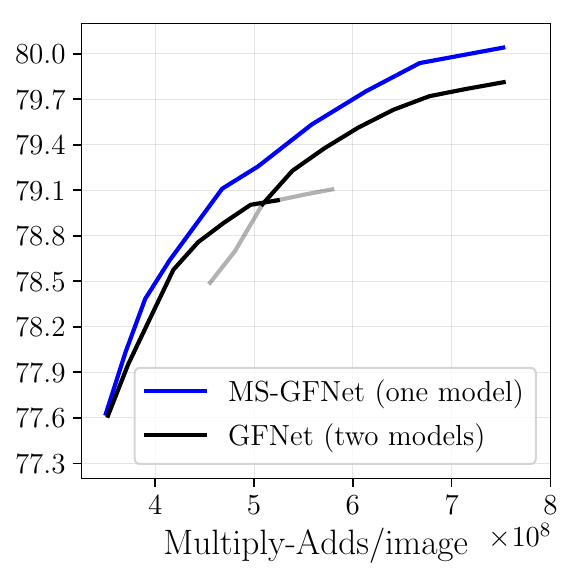}
      \end{minipage}}
      \vskip -0.2in
      \caption{\label{fig:MS_vs_original} Comparisons of MS-GFNet and the original GFNet on top of ResNet and EfficientNet. Top-1 accuracy v.s. Multiply-Adds on ImageNet under the \textit{Budgeted batch classification} setting are reported.}
    \end{minipage}
    \vskip -0.15in
\end{figure}

\subsubsection{MS-GFNet}
\label{sec:exp_msgf}

The performance of MS-GFNet is presented in Table \ref{tab:MS-GFNet}. The proposed model is compared with the vanilla GFNets that adopt the same patch size as it in the \emph{Glance Step} (\emph{i.e.}, 96x96), but use a fixed patch size in the \emph{Focus Stage} (\emph{i.e.}, 96x96, 128x128, 160x160). All the models in Table \ref{tab:MS-GFNet} is trained with the aforementioned contrastive reward. One can observe that smaller patches outperforms larger ones when the computational budget is small, while larger patches achieve higher accuracy on the contrary. In contrast, MS-GFNet consistently outperforms the best results of fixed-patch-size baselines with all computational budgets, which demonstrates the effectiveness of the proposed multi-scale patch mechanism. Note that MS-GFNet has the same network architecture with the baselines except for the slight changes in $\pi$, and the number of parameters is approximately identical.

In Figure \ref{fig:MS_vs_original}, we compare the vanilla GFNet and the MS-GFNet trained with the contrastive reward. The latter not only consistently outperforms the former under the same computational budget, but is able to achieve higher efficiency among a larger range of Multiply-Adds. This enables MS-GFNet to adjust its computational cost more flexibly without additional training in realistic scenarios.



\begin{figure}[!t]
    \begin{center}
    \centerline{\includegraphics[width=0.75\columnwidth]{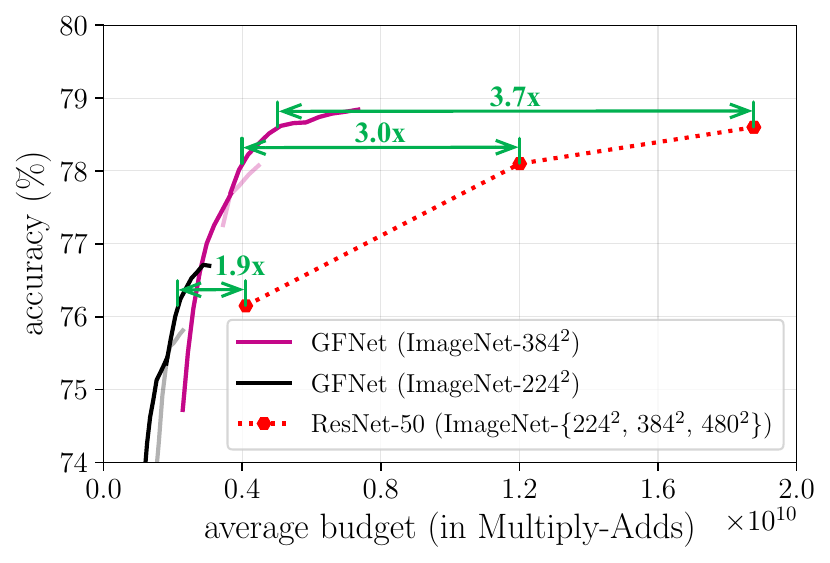}}
    \vspace{-2.25ex}
    \caption{Top-1 accuracy v.s. Multiply-Adds on ImageNet with varying input resolutions. The images are pre-processed by being resized to $224^2$, $384^2$ or $480^2$. Our GFNet is implemented on top of ResNet-50.
    }
    \label{fig:imagenet_high_res}
    \end{center}
    \vspace{-2.5ex}
\end{figure}

\begin{table}[t]
	\footnotesize
    \centering
    \caption{Results of traffic sign recognition on the Swedish traffic signs dataset, which consists of 960x1,280 road-scene images collected on real moving vehicles. The patch size for GFNet is set to $H'\!=\!W'\!=\!192$, while ``random patch'' denotes randomly cropping the patches in our method. The best results are \textbf{bold-faced}. The \textcolor{blue}{\textbf{blue}} numbers are based on the backbone network (\emph{i.e.}, ResNet-18). $^\dagger$The computational cost of GFNet can be adjusted online.}
    \vskip -0.15in
    \label{tab:traffic_sign}
    \setlength{\tabcolsep}{2.5mm}{
    \vspace{5pt}
    \renewcommand\arraystretch{1.2} 
    \begin{tabular}{|c|cc|}
        \hline
        \multirow{2}{*}{Model} & Multiply-Adds & \multirow{2}{*}{Top-1 Accuracy} \\
        & per image & \\
    \hline
    ResNet-18 & 44.7G & 89.47\% \\
    GF-ResNet-18 & \multirow{2}{*}{2.8G} & \multirow{2}{*}{68.97\%} \\[-0.025in]
    (random patch) && \\
    \hline
    \multirow{2}{*}{GF-ResNet-18$^\dagger$}  & \textbf{1.9G}$_{\bm{{\textcolor{blue}{\downarrow23.5\times}}}}$ & 89.47\% \\
    & 2.8G$_{{{\textcolor{blue}{\downarrow16.0\times}}}}$ & \textbf{91.50\%} \\
    \hline
    \end{tabular}}
    \vskip -0.15in
\end{table}

\begin{figure*}[t]
    \begin{center}
    \centerline{\includegraphics[width=2.0\columnwidth]{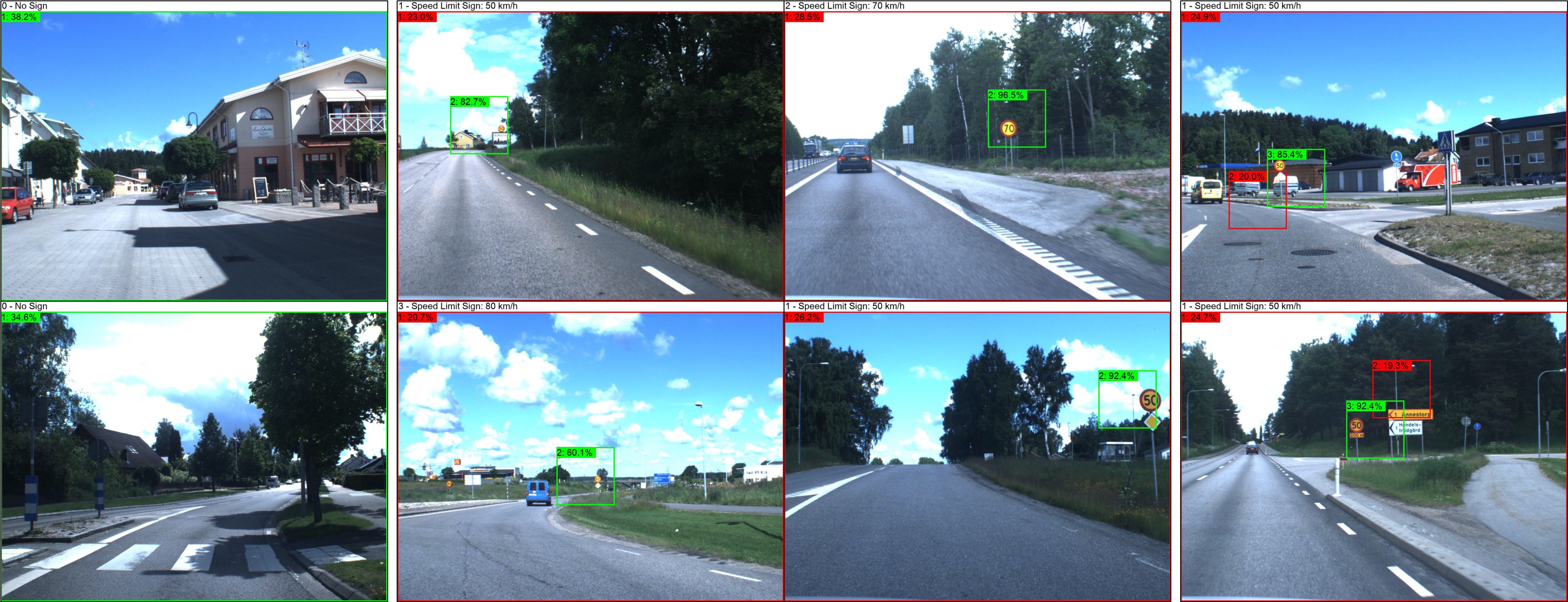}}
    \vspace{-1.5ex}
    \caption{Visualization of traffic sign recognition. The boxes indicate the patch locations, while the color denotes whether the prediction is correct at the current step (green: correct; red: wrong). 
    }
    \label{fig:visualization_traffic_sign}
    \end{center}
    \vspace{-1ex}
\end{figure*}

\begin{figure*}[!t]
    \centering
    \vspace{-2.5ex}
    \subfigure[Something-Something V1]{\hspace{-0.1in}
    \begin{minipage}[t]{0.35\linewidth}
    \centering
    \includegraphics[width=\linewidth]{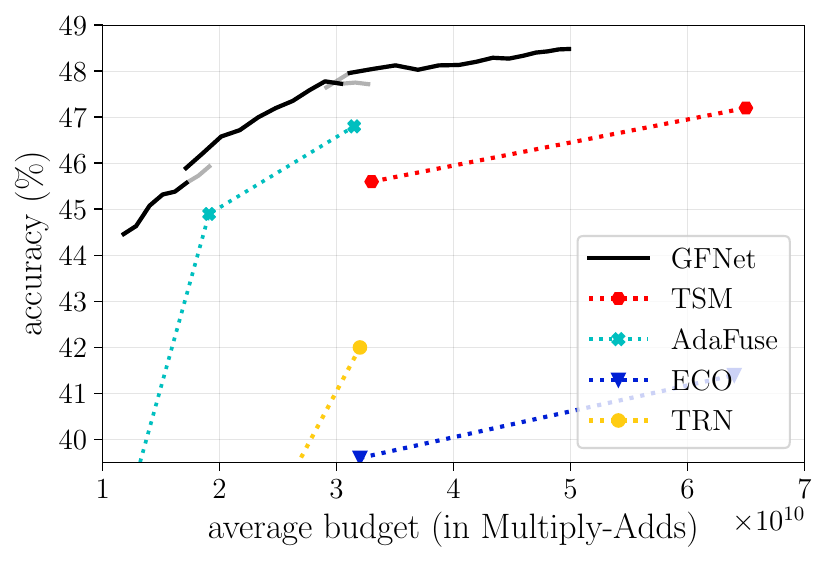}
    \end{minipage}%
    }\hspace{-0.1in}
    \subfigure[Something-Something V2]{
    \begin{minipage}[t]{0.33\linewidth}
    \centering
    \includegraphics[width=\linewidth]{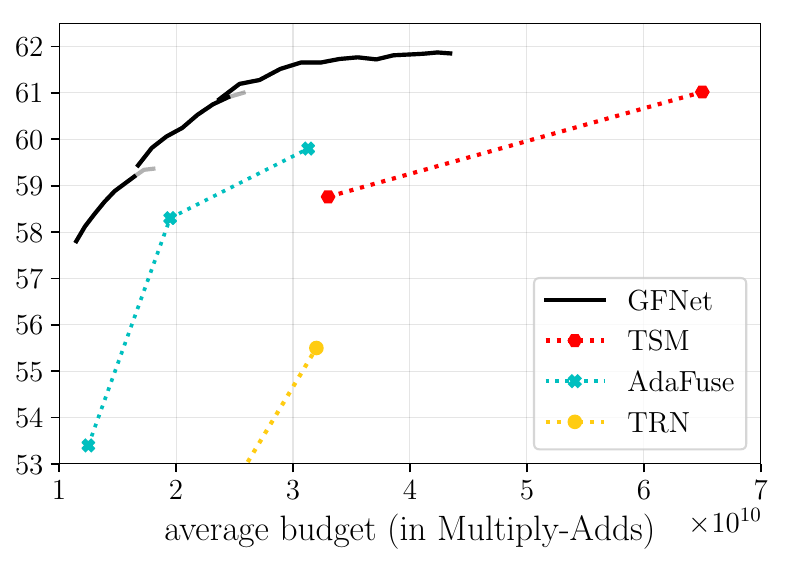}
    \end{minipage}%
    }\hspace{-0.1in}
    \subfigure[Jester]{
    \begin{minipage}[t]{0.33\linewidth}
    \centering
    \includegraphics[width=\linewidth]{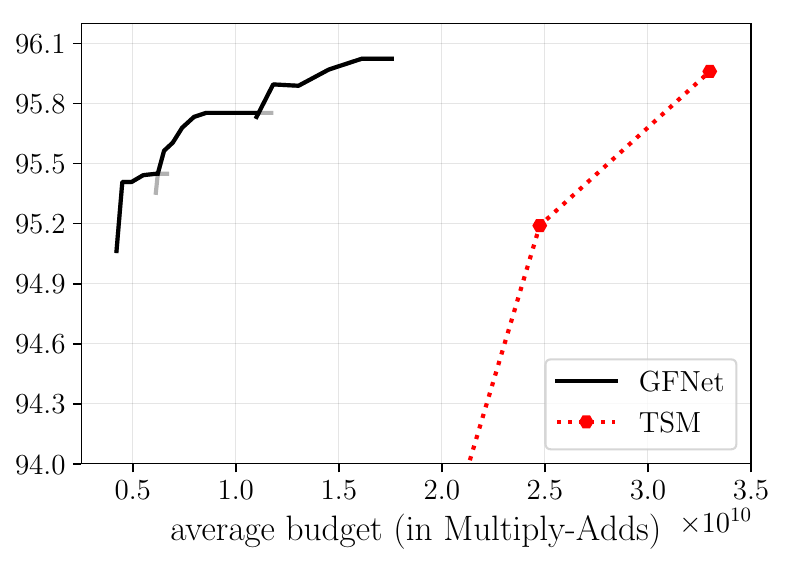}
    \end{minipage}%
    }%
    \centering
    \vspace{-2ex}
    \caption{\label{fig:video_recognition}Top-1 accuracy v.s. Multiply-Adds on three video recognition benchmarks under the \textit{Budgeted batch classification} setting. GFNet is implemented on top of a ResNet-50 with the temporal shift module (TSM) \cite{lin2019tsm}.}
    \vspace{-1ex}
\end{figure*}

\subsection{High-resolution Image Recognition}
\label{sec:high_res}

\textbf{Results on ImageNet with higher resolution.}
As aforementioned, leveraging high-resolution inputs significantly improves the accuracy of modern deep networks. However, large images usually yield a high computational cost, which grows quadratically or sometimes even faster with respect to the image height (or width). One of the predominant advantages of our method is that GFNet can process high-resolution inputs efficiently, while preserving their gains in accuracy. To demonstrate this point, we pre-process the ImageNet dataset to obtain the images with varying resolutions (\emph{i.e.}, $224^2$, $384^2$ and $480^2$), and compare GFNet with the baselines in these scenarios, as shown in Figure \ref{fig:imagenet_high_res}. The patch size of GFNet is set to 96x96/128x128 and 160x160/192x192 for the original image size of $224^2$ and $384^2$, respectively. The results in Figure \ref{fig:imagenet_high_res} indicate that our method outperforms the backbone networks by increasingly large margins with the growing resolution of original images. For example, on top of the same $224^2$/$384^2$ images, GF-ResNet-50 reduces the computational cost of ResNet-50 by $1.9\times$/$3.0\times$ without sacrificing the accuracy. Another interesting phenomenon is that GFNet achieves higher best accuracies than the backbones when leveraging the same inputs. We tentatively attribute this to the paradigm of dynamic computation, which allocates more computation to the task-relevant regions adaptively and may learn more discriminative representations.

\textbf{Traffic sign recognition.}
As a standard benchmark, the ImageNet dataset mainly contains the images that have already been centered to the relevant object by the human photographers. However, our GFNet does not rely on this assumption, and is applicable to more general scenarios, \emph{e.g.}, where the test images may be collected in the wild without specified pre-processing. As a representative example, we present the results on the Swedish traffic signs dataset in Table \ref{tab:traffic_sign}. The dataset consists of 960x1,280 road-scene images collected on real moving vehicles, and the task is to recognize the existence and types of the speed limit signs. Note that the objects of interest are generally small, diversely distributed, and sometimes not clear (see Figure \ref{fig:visualization_traffic_sign} for examples). From Table \ref{tab:traffic_sign}, one can observe that GFNet dramatically improves the computational efficiency of the backbone network, \emph{e.g.}, it reduces the computational cost by $23.5\times$ with the preserved accuracy. Several representative visualization examples are presented in Figure \ref{fig:visualization_traffic_sign}. We find that GFNet is able to identify the existence of traffic signs effectively at the \textit{Glance Step}, and can further attend to the local regions that contain the signs to recognize their specific contents, in the \textit{Focus Stage}. In addition, once GFNet fails to localize the traffic signs of interest (\emph{e.g.}, due to other distracting signs), it may simply correct this with additional focus steps.


\begin{figure*}[!t]
    \begin{center}
    \centerline{\includegraphics[width=2\columnwidth]{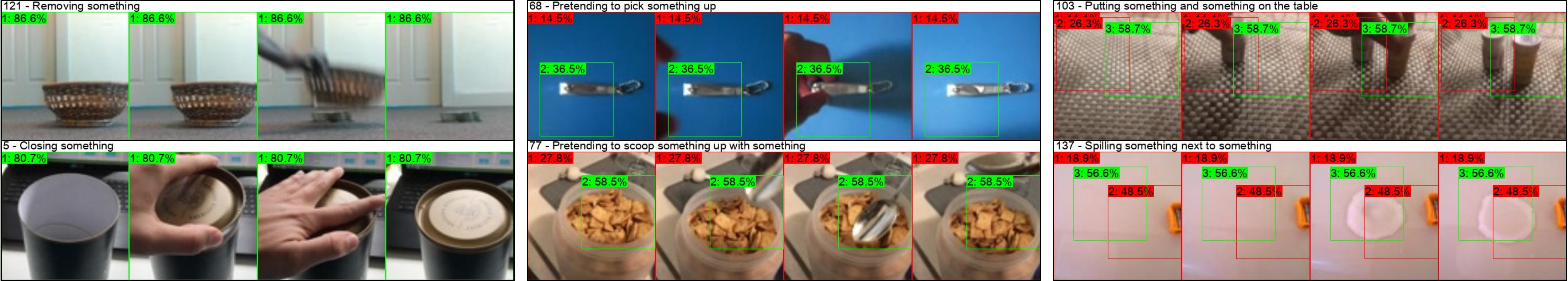}}
    \vspace{-1.5ex}
    \caption{Visualization of the GFNet on the representative videos from Something-Something V2. The boxes indicate the patch locations, and the color denotes whether the prediction is correct at current step (green: correct; red: wrong). Note that $\tilde{\bm{x}}_1$ is the resized input video. The indices of the steps and the current confidence on the ground truth labels (shown at the top of images) are presented in the upper left corners of boxes. 
    }
    \label{fig:video_visualization}
    \end{center}
    \vspace{-1ex}
\end{figure*}

\subsection{Video Recognition}
\label{sec:video_recognition}

\textbf{Main results.}
The \textit{Budgeted batch classification} results on three representative video-based benchmarks are shown in Figure \ref{fig:video_recognition}, where similar observations to image recognition can be obtained. GFNet is shown to significantly improve the computational efficiency of TSM. For instance, on Something-Something V2, GFNet reduces the average budget for each video from $6.5\times10^{10}$ to $2.5\times10^{10}$ Multiply-Adds when reaching an accuracy of $\sim61\%$. In addition, GFNet outperforms the state-of-the-art efficient video recognition framework, AdaFuse \cite{meng2021adafuse}, by large margins ($\sim1.5-5\%$). Notably, however, GFNet is actually orthogonal to AdaFuse since the latter mainly focuses on improving the network architecture. Besides, in video recognition, the computational cost of our method can also be tuned online without additional training.

\textbf{Visualization.}
The visualization results of GFNet on video benchmarks are shown in Figure \ref{fig:video_visualization}. For each video, 4 uniformly-sampled frames are presented, and the meaning of annotations is the same as Figure \ref{fig:visualization}. One can observe that GFNet is able to recognize some semantically simpler actions at the \textit{Glance Step}, \emph{e.g.}, ``removing something'', which mainly includes identifying the disappearance of a certain object. In contrast, for more difficult samples, our model can adaptively localize and leverage some task-relevant video patches to understand more complex behaviors (\emph{e.g.}, pretending to do something) or model the relationships between objects (\emph{e.g.}, the co-occurrence of two objects and the ``next to'' relationship).


\begin{figure*}
    \vspace{-0.025in}
    \begin{minipage}[t]{0.32\linewidth}
        \centering
        \includegraphics[width=1\textwidth]{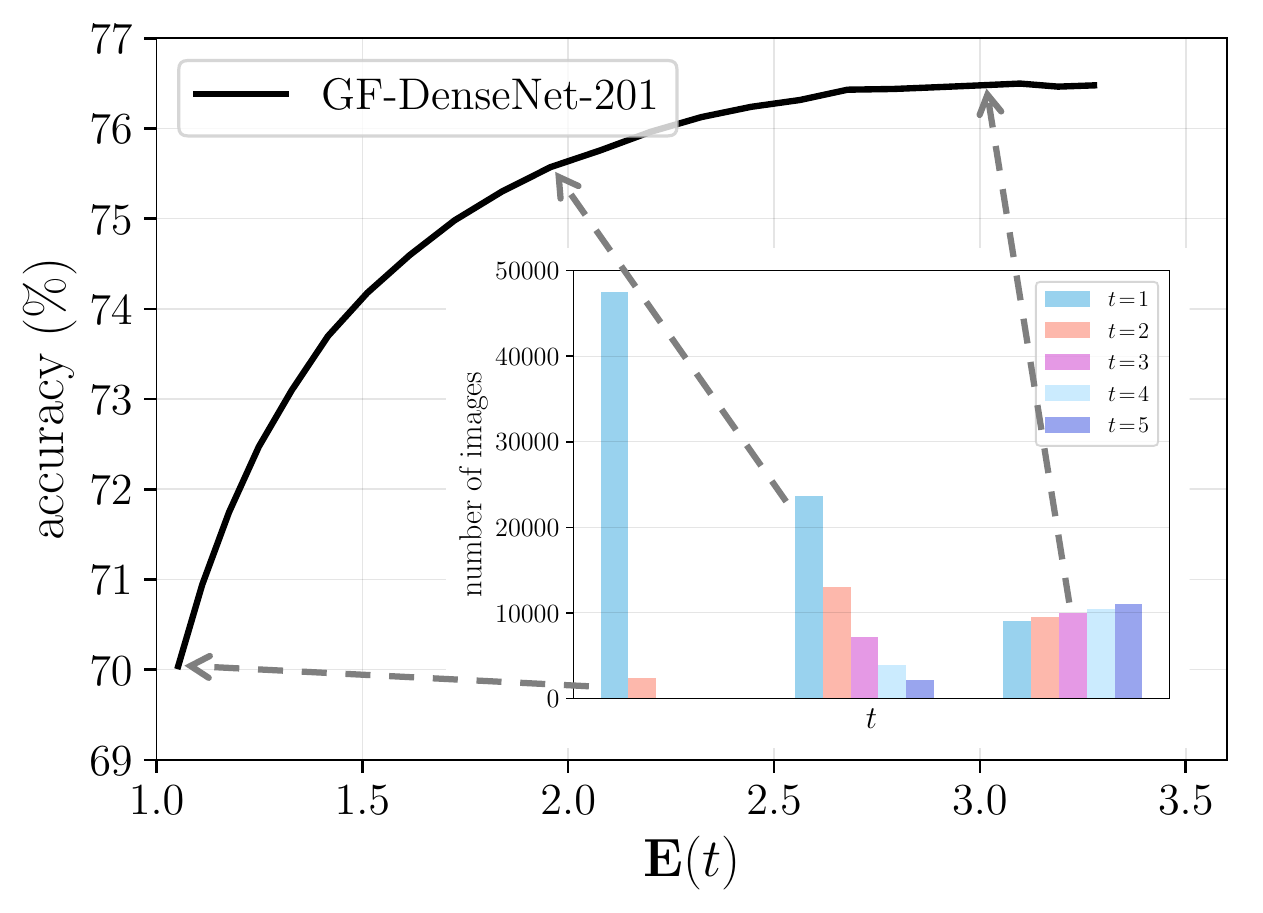}
        \vskip -0.15in
        \caption{\label{fig:e_t}Top-1 accuracy v.s. the expected input sequence length during inference, \textbf{\textnormal{E}}($t$), in \textit{Budgeted batch classification}. The results are based on GF-DenseNet-201 ($T\!\!=\!\!5$, $H'\!\!=\!\!W'\!\!=\!96$).}
    \end{minipage}
    \hspace{0.02in}
    \begin{minipage}[t]{0.339\linewidth}
        \centering
        \includegraphics[width=1\textwidth]{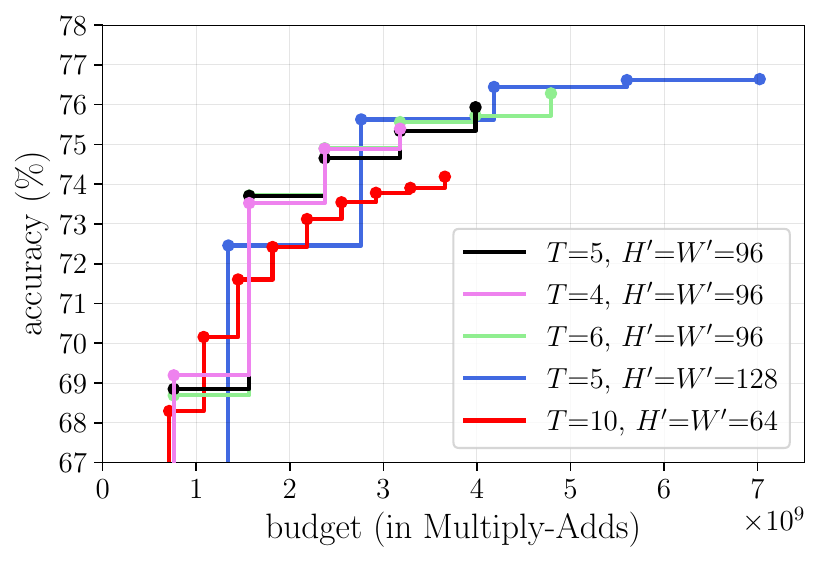}
        \vskip -0.15in
        \caption{\label{fig:T_WH}Performance of GFNet with varying $T$ and different patch sizes. Here we use ResNet-50 as the two backbones. The \textit{Anytime prediction} results are reported.}
    \end{minipage}
    \hspace{0.02in}
    \begin{minipage}[t]{0.32\linewidth}
        \centering
        \includegraphics[width=1\textwidth]{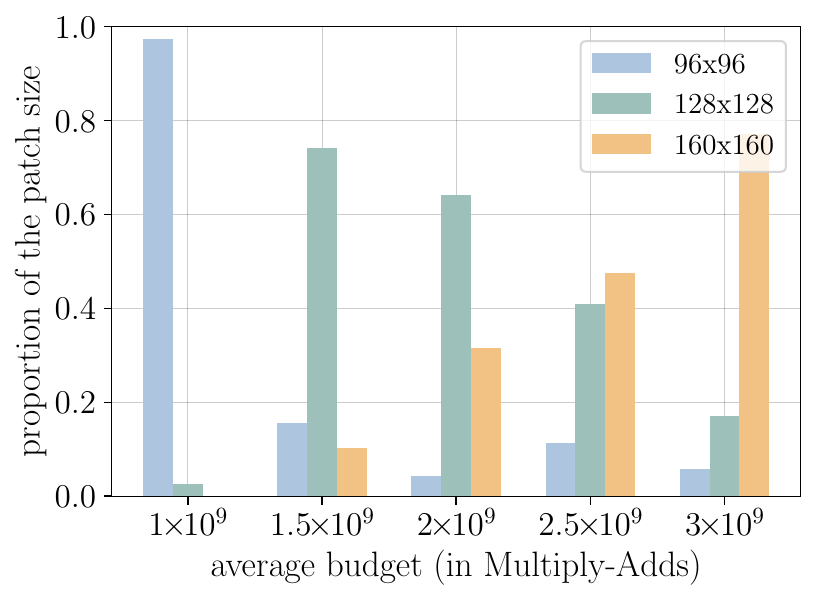}
        \vskip -0.15in
        \caption{\label{fig:gfv2_ms_ratio}The proportion of each patch size adopted in an MS-GFNet (ResNet-50) with varying computational budget. The \textit{Budgeted batch classification} setting is considered.}
    \end{minipage}
    \vskip -0.1in
\end{figure*}

 \begin{table*}[!t]
	\footnotesize
    \centering
    \caption{Ablation study on the components of GFNet. We report the Top-1 accuracy of different variants with fixed length of the input sequence (denoted by $t$). The best results are \textbf{bold-faced}.}
    \vskip -0.15in
    \label{tab:abl}
    \setlength{\tabcolsep}{2mm}{
    \vspace{5pt}
    \renewcommand\arraystretch{1.25} 
    \begin{tabular}{|c|l|ccccc|}
        \hline
    \multicolumn{2}{|c|}{Variants} & $t=1$ & $t=2$ & $t=3$ & $t=4$ & $t=5$ \\
    \hline
    \multirow{4}{*}{\shortstack{Patch Selection}}& Random Policy &54.53\%&64.42\%&68.09\%&69.85\%&70.70\% \\
    &Random Policy + \textit{Glance Step}&69.47\%&72.32\%&73.47\%&74.04\%&74.46\% \\
    &Centre-corner Policy &59.51\%&66.75\%&69.53\%&72.83\%&74.10\% \\
    &Centre-corner Policy + \textit{Glance Step} &69.06\%&72.94\%&73.88\%&74.47\%&75.12\% \\
    \hline
    \multirow{2}{*}{\shortstack{Network}} & w/o Global Encoder $f_{\textsc{g}}$ (single encoder) &65.92\%&70.59\%&72.59\%&73.72\%&74.26\% \\
    & w/o the \textit{Glance Step} (using a centre crop as $\tilde{\bm{x}}_1$) &59.14\%&66.70\%&70.91\%&73.71\%& 74.70\%\\
    \hline
    \multirow{2}{*}{\shortstack{Training}} 
    &Initializing $f_{\textsc{g}}$ w/o $H'\!\times\!W'$ Fine-tuning &67.81\%&72.50\%&74.12\%&75.21\%&75.56\% \\
    &w/o Training Stage \uppercase\expandafter{\romannumeral3} (2-stage training) &\textbf{69.62\%}&73.39\%&74.32\%&74.97\%&75.36\% \\
    \hline
    \multicolumn{2}{|c|}{GFNet (ResNet-50, $T=5$, $H'\!=\!W'\!=\!96$)} &68.85\%&\textbf{73.71\%}&\textbf{74.65\%}&\textbf{75.34\%}&\textbf{75.93\%} \\
    \hline
    \end{tabular}}
    \vskip -0.1in
\end{table*}

\begin{table*}[!t]
	\footnotesize
    \centering
    \caption{Comparisons of different architectures of the patch proposal network $\pi$. Here ``Conv.'' denotes the convolutional layer. We report the ImageNet Top-1 accuracy of GF-ResNet-50 ($T\!\!=\!\!5$, $H'\!\!=\!\!W'\!\!=\!96$) with fixed length of the input sequence (denoted by $t$). For clear comparisons, we provide the computational cost of representative backbone networks (with 96x96 inputs) in the last two rows. Our proposed efficient policy network introduces negligible additional computation on top of the light-weighted backbones, but performs on par with the computationally intensive designs.}
    \vskip -0.15in
    \label{tab:arch_pi}
    \setlength{\tabcolsep}{2mm}{
    \vspace{5pt}
    \renewcommand\arraystretch{1.25} 
    \begin{tabular}{|cc|c|ccccc|}
        \hline
    \multicolumn{2}{|c|}{Architecture of $\pi$.} & \multirow{2}{*}{\shortstack{Multiply-Adds\\per Step}} & \multirow{2}{*}{$t=1$} & \multirow{2}{*}{$t=2$} & \multirow{2}{*}{$t=3$} & \multirow{2}{*}{$t=4$} & \multirow{2}{*}{$t=5$} \\
    Head (before GRU) & GRU Hidden Units &&&&&&\\
    \hline
    Fully-connected Layer & 1,024 & 46.1M & 68.85\% & {73.71\%} & {74.65\%} & {75.34\%} & {75.93\%} \\
    Fully-connected Layer & 512  & 40.4M & 69.09\% & 73.81\% & 74.83\% & 75.39\% & 75.83\% \\
    Fully-connected Layer & 256  & 38.7M & 68.88\% & 73.63\% & 74.81\% & 75.25\% & 75.67\% \\
    3x3 Conv. $\to$ 32 Channels & 256  & 5.8M & 68.89\% & 73.65\% & 74.81\% & 75.54\% & 75.75\% \\
    1x1 Conv. $\to$ 64 Channels & 256  & 1.7M & 68.77\% & 73.71\% & 74.90\% & 75.39\% & 75.87\% \\
    1x1 Conv. $\to$ 32 Channels & 256  & 1.1M & 68.75\% & 73.77\% & 74.99\% & 75.54\% & 75.86\% \\
    1x1 Conv. $\to$ 16 Channels & 256  & \textbf{0.7M} & 68.57\% & 73.78\% & 74.95\% & 75.58\% & 75.85\% \\
    \hline
    \multicolumn{2}{|c|}{ResNet-50, Input Size: 96x96} & 750.7M & \multicolumn{5}{c|}{--} \\
    \multicolumn{2}{|c|}{MobileNet-V3-Large (1.00), Input Size: 96x96} & 43.1M & \multicolumn{5}{c|}{--} \\
    \hline
    \end{tabular}}
    \vskip -0.1in
\end{table*}

\subsection{Analytical Results}
\label{subsec:ablation}


\subsubsection{Input Sequence Length and Patch Size}

\textbf{Length of input sequence.}
To give a deeper understanding of our method, we visualize the expected length of the input sequence \textbf{\textnormal{E}}($t$) during inference under the \textit{Budgeted batch classification} setting and the corresponding Top-1 accuracy on ImageNet, as shown in Figure \ref{fig:e_t}. We also present the plots of numbers of images v.s. $t$ at several points. It can be observed that the performance of GFNet is significantly improved by letting images exit later in the \textit{Focus Stage}, which is achieved by adjusting the confidence thresholds online without additional training.

\textbf{Maximum input sequence length $T$ and patch size.}
We show the performance of GFNet with varying $T$ and different patch sizes under the \textit{Anytime prediction} setting in Figure \ref{fig:T_WH}. The figure suggests that changing $T$ does not significantly affect the performance with the same amount of computation, while using larger patches leads to better performance with large computational budgets but lower test accuracy compared with smaller patches when the computational budget is insufficient.

\textbf{Patch sizes in MS-GFNet.}
Figure \ref{fig:gfv2_ms_ratio} visualizes the proportion of each patch size candidate among all focus patches in the MS-GFNet (ResNet-50) in Table \ref{tab:MS-GFNet}. Here we omit the \emph{Glance Step} as it uses a fixed patch size. The results show that MS-GFNet is able to switch between adopting small and large patches when limited and relatively sufficient computational resources are available.

\begin{table}[t]
	\footnotesize
    \centering
    \caption{Comparisons of various early-termination criterions in \textit{Budgeted batch classification}. Results with a GF-ResNet-50 ($T\!\!=\!\!5$, $H'\!\!=\!\!W'\!\!=\!96$) are presented. The best Top-1 accuracy with each computational budget are \textbf{bold-faced}.}
    \vskip -0.1in
    \label{tab:abl_early_criterion}
    \setlength{\tabcolsep}{0.35mm}{
    \vspace{5pt}
    \renewcommand\arraystretch{1.35} 
    \begin{tabular}{|c|lllll|}
        \hline
    Early-exit & \multicolumn{5}{c|}{Average Budget (in Multiply-Adds)} \\
    Criterion &  \ 1.00G & 1.25G & 1.50G & 1.75G & 2.00G  \\
    \hline
    Random &\ 69.99\%&70.89\%&71.68\%&72.17\%&72.87\% \\
    Entropy &\ 72.20\%&74.01\%&74.75\%&75.19\%&75.43\% \\
    Confidence &\ \textbf{72.64\%}{\tiny{$\pm$0.08}} &\textbf{74.38\%}{\tiny{$\pm$0.09}} &\textbf{75.12\%}{\tiny{$\pm$0.05}} &\textbf{75.46\%}{\tiny{$\pm$0.06}} &\textbf{75.67\%}{\tiny{$\pm$0.03}}  \\
    \hline
    \end{tabular}}
\end{table}

\begin{table}[t]
	\footnotesize
    \centering
    \caption{Effects of varying training hyper-parameters, where $lr$ refers to the learning rate. The results of GF-ResNet-50 ($T\!\!=\!\!5$, $H'\!\!=\!\!W'\!\!=\!96$) are presented as the representative examples. We report the ImageNet Top-1 accuracy with fixed length of the input sequence (denoted by $t$). The best results are \textbf{bold-faced}. Our adopted settings are \underline{underlined}.} 
    \vskip -0.1in
    \label{tab:sensitivity}
    \setlength{\tabcolsep}{1.4mm}{
    \vspace{5pt}
    \renewcommand\arraystretch{1.35} 
    \begin{tabular}{|l|ccccc|}
        \hline
    Hyper-parameters & $t=1$ & $t=2$ & $t=3$ & $t=4$ & $t=5$ \\
    \hline
    $\lambda=0$ (\emph{i.e.}, Eq. (2)) &68.55\%&71.87\%&73.19\%&73.52\%&73.94\% \\
    \underline{$\lambda=1$ (ours)} &\underline{68.85\%}&\underline{{73.71\%}}&\underline{\textbf{74.65\%}}&\underline{\textbf{75.34\%}}&\underline{\textbf{75.93\%}} \\
    $\lambda=2$ & 69.25\% & 73.95\% & 74.32\% & 75.19\% & 75.73\% \\
    $\lambda=5$ & 69.39\% & \textbf{73.98\%} & 74.10\% & 74.51\% & 74.77\%  \\
    $\lambda=10$ & \textbf{69.71\%} & 73.71\% & 73.84\% & 74.35\% & 74.70\%  \\
    \hline
    Initial $lr$ = 0.001 & 68.75\% & 72.20\% & 73.44\% & 74.18\% & 74.47\% \\
    \underline{Initial $lr$ = 0.01 (ours)} & \underline{\textbf{68.85\%}} &\underline{\textbf{73.71\%}}&\underline{74.65\%}&\underline{\textbf{75.34\%}}&\underline{\textbf{75.93\%}} \\
    Initial $lr$ = 0.1 & 68.17\% & 73.58\% & \textbf{74.78\%} & 75.00\% & 75.83\% \\
    \hline
    30 Epochs & \textbf{69.02\%} & 73.45\% & \textbf{74.74\%} & 75.25\% & 75.55\% \\
    \underline{60 Epochs (ours)} &\underline{68.85\%}&\underline{{73.71\%}}&\underline{{74.65\%}}&\underline{\textbf{75.34\%}}&\underline{\textbf{75.93\%}} \\
    90 Epochs & 68.43\% & \textbf{74.01\%} & 74.49\% & 75.23\% & 75.85\% \\
    \hline
    \end{tabular}}
\end{table}

\subsubsection{Ablation Study}
\label{sec:abl_study}

\textbf{Effectiveness of the components.}
We test ablating the components of GFNet, and summarize the results in Table \ref{tab:abl}. We first consider two alternatives of the learned patch selection policy, namely the random policy where all patches are uniformly sampled from the image, and the centre-corner policy where the network sequentially processes the whole image by first cropping the patch from the centre of the image, and then traversing the corners. The learned policy is shown to consistently outperform them. We also remove or alter the components of the GFNet architecture and the training process. One can observe that resizing the original image as $\tilde{\bm{x}}_1$ (\textit{Glance Step}) and adopting two encoders are both important techniques to achieve high accuracy, especially at the first three steps.

\textbf{Architecture of the patch proposal network $\pi$.} 
Our proposed GFNet is robust to the architecture of $\pi$. In particular, since $\pi$ receives the discriminative deep representations extracted by the two encoders, a high network capacity is generally not necessary to process its inputs. In Table \ref{tab:arch_pi}, we compare the performance of various architectures of $\pi$. The results suggest that a lighted-weighted convolutional layer with a small number of GRU hidden units achieves similar accuracy to the computationally more expensive designs of $\pi$. Our proposed architecture (\emph{i.e.}, Figure \ref{fig:PPN}) introduces negligible additional computation compared to the backbone networks (\emph{e.g.}, ResNet-50 and MobileNet-V3).


\textbf{Early-termination criterion.}
We change the criterion for performing adaptive inference, and report the results in Table \ref{tab:abl_early_criterion}. Two variants are considered: (1) performing random termination at each step with the same proportion as GFNet; (2) adopting the entropy of the softmax prediction to determine whether to terminate the inference process. One can observe that our simple but effective confidence-based criterion consistently outperforms both of them.

\subsubsection{Sensitivity of Training Hyper-parameters}
\label{sec:hyper_sense}

In our experiments, three training hyper-parameters are tuned conditioned on the backbone networks, while other training configurations introduced by GFNet are fixed (\emph{e.g.}, reinforcement learning). In specific, the tunable hyper-parameters include: (1) the coefficient $\lambda$ in the loss function (Eq. (\ref{eq:loss_cls_impl})); (2) the initial learning rate for the backbone networks in training stage \uppercase\expandafter{\romannumeral1}/\uppercase\expandafter{\romannumeral3}; (3) the number of epochs in training stage \uppercase\expandafter{\romannumeral1}/\uppercase\expandafter{\romannumeral3}. We study their sensitivity in Table \ref{tab:sensitivity}. 

One can observe that a proper $\lambda$ significantly improves the accuracy of GFNet. However, too large $\lambda$ moderately hurts the final performance with a relatively long input sequence, in which cases the gradients from the classifier $f_{\textsc{c}}$ may be overwhelmed in the two encoders. For the initial learning rate of training stage \uppercase\expandafter{\romannumeral1}/\uppercase\expandafter{\romannumeral3}, we find that it can be straightforwardly set to 1/10 of the values for training the corresponding backbones from scratch (\emph{e.g.}, 0.01 for ResNet-50). In addition, as shown in Table \ref{tab:sensitivity}, 60-90 training epochs are generally sufficient to achieve the saturated performance of GFNet on ImageNet.

\textbf{Applying GFNet to new backbones/tasks.}
In general, when deploying GFNet in new scenarios, one can directly adopt most of our proposed training settings (\emph{e.g.}, with our released code or pre-trained models), and configure the fine-tuning learning rate following the aforementioned rule. The hyper-parameter searching may be performed on $\lambda$ and the number of training epochs, where $\lambda=1$ and 60 epochs can be a good starting point or a preliminary setting for the straightforward implementation.











%% file: conclusion.tex

\section{Conclusion}
In this paper, we introduced a \emph{Glance and Focus Network} (GFNet) to reduce the spatial redundancy in image and video recognition tasks. GFNet processes a given high-resolution input in a sequential manner. At each step, GFNet processes a smaller input, which is either a down-sampled version of the original image/video or a cropped patch. GFNet progressively performs classification as well as localizing discriminative regions for the next step. This procedure is terminated once sufficient classification confidence is obtained, leading to an adaptive inference paradigm.
Our method is compatible with a wide variety of visual backbones and is easy to implement on mobile devices and GPUs. Extensive experiments on four large-scale benchmarks showed that GFNet significantly improves the computational efficiency even on top of the most state-of-the-art light-weighted CNNs. Future work may focus on extending the framework to downstream tasks such as object detection and semantic segmentation. It is also interesting to explore the automatic patch localization ability of GFNet for weakly supervised object localization or detection.




%% file: appendix.tex
\appendices

\section{Implementation Details}
\subsection{Recurrent Networks}
\label{app:recurrent_net}
For RegNets \cite{radosavovic2020designing}, MobileNets-V3 \cite{howard2019searching} and EfficientNets \cite{DBLP:conf/icml/TanL19}, we use a gated recurrent unit (GRU) with 256 hidden units \cite{cho2014learning} in the patch proposal network $\pi$. For ResNets \cite{He_2016_CVPR} and DenseNets \cite{2019arXiv160806993H}, we adopt 1024 hidden units and remove the convolutional layer in $\pi$. This does not hurt the efficiency since here the computational cost of $\pi$ is negligible compared with the two encoders. With regards to the recurrent classifier $f_{\textsc{c}}$, for ResNets \cite{He_2016_CVPR}, DenseNets \cite{2019arXiv160806993H} and RegNets \cite{radosavovic2020designing}, we use a GRU with 1024 hidden units. For MobileNets-V3 \cite{howard2019searching} and EfficientNets \cite{DBLP:conf/icml/TanL19}, we find that although a GRU classifier with a large number of hidden units achieves excellent classification accuracy, it is excessively computationally expensive in terms of efficiency. Therefore, we replace the GRU with a cascade of fully connected classification layers. In specific, at $t^{\textnormal{th}}$ step, we concatenate the feature vectors of all previous inputs $\{\bm{\bar{e}}_1, \dots, \bm{\bar{e}}_{t}\}$, and use a linear classifier with the size $tF\!\times\!C$ for classification, where $F$ is the number of feature dimensions and $C$ is the number of classes. Similarly, we use another $(t\!+\!1)F\!\times\!C$ linear classifier at $(t\!+\!1)^{\textnormal{th}}$ step. Totally, we have $T$ linear classifiers with the size $F\!\times\!C, 2F\!\times\!C, \dots, TF\!\times\!C$.


\subsection{Policy Gradient Algorithm}
\label{app:PPO}
During training, the objective of the patch proposal network $\pi$ is to maximize the sum of the discounted rewards:
\begin{equation}
    \label{eq:app_reward}
    \max_{\pi} \mathbb{E}\left[\sum\nolimits_{t=2}^{T} \gamma^{t-2} r_t \right],
\end{equation}
where, $\gamma \in (0,1)$ is a pre-defined discount factor, $r_{t}$ is the reward for the localization action $\bm{l}_{t}$, and $T$ is the maximum length of the input sequence. The action $\bm{l}_{t}$ is stochastically chosen from a distribution parameterized by $\pi$: $\bm{l}_{t} \sim \pi(\bm{l}_{t}|\bm{e}_{t-1}, \bm{h}^{\pi}_{t-2})$, where we denote the hidden state maintained within $\pi$ by $\bm{h}^{\pi}_{t-2}$. Here we use a Gaussian distribution during training, whose mean is outputted by $\pi$ and standard deviation is pre-defined as a hyper-parameter. At test time, we simply adopt the mean value as $\bm{l}_{t}$ for a deterministic inference process. Note that, we always resize the original image $\bm{x}$ to $H'\!\times\!W'$ as $\tilde{\bm{x}}_1$ (\textit{Glance Step}), and thus we do not have $\bm{l}_{1}$ or $r_1$. 

In this work, we implement the proximal policy optimization (PPO) algorithm proposed by \cite{schulman2017proximal} to train the patch proposal network $\pi$. In the following, we briefly introduce its procedure. For simplicity, we denote $\pi(\bm{l}_{t}|\bm{e}_{t-1}, \bm{h}^{\pi}_{t-2})$ by $\pi(\bm{l}_{t}|\bm{s}_t)$, where $\bm{s}_t$ is the current state containing $\bm{e}_{t-1}$ and $\bm{h}^{\pi}_{t-2}$. First, we consider a surrogate objective:
\begin{equation}
    {L}^{\textnormal{CPI}}_t = \frac{\pi(\bm{l}_{t}|\bm{s}_t)}{\pi_{\textnormal{old}}(\bm{l}_{t}|\bm{s}_t)} \hat{A}_t,
\end{equation}
where $\pi_{\textnormal{old}}$ and $\pi$ are the patch proposal network before and after the update, respectively. The advantage estimator $\hat{A}_t$ is computed by:
\begin{equation}
    \hat{A}_t = -V(\bm{s}_t) + r_t + \gamma r_{t+1} + \dots + \gamma^{T-t} r_T,
\end{equation}
where $V(\bm{s}_t)$ is a learned state-value function that shares parameters with the policy function (they merely differ in the final fully connected layer). Since directly maximizing ${L}^{\textnormal{CPI}}$ usually leads to an excessively large policy update, a clipped surrogate objective is adopted \cite{schulman2017proximal}:
\begin{equation}
    {L}^{\textnormal{CLIP}}_t =  
        \textnormal{min} \left\{
            \frac{\pi(\bm{l}_{t}|\bm{s}_t)}{\pi_{\textnormal{old}}(\bm{l}_{t}|\bm{s}_t)} \hat{A}_t,
            \textnormal{clip}(
                \frac{\pi(\bm{l}_{t}|\bm{s}_t)}{\pi_{\textnormal{old}}(\bm{l}_{t}|\bm{s}_t)}, 1-\epsilon, 1+ \epsilon
            )\hat{A}_t
        \right\},
\end{equation}
where $0 < \epsilon < 1$ is a hyper-parameter. Then we are ready to give the final maximization objective:
\begin{equation}
    \label{eq:objective}
    \mathop{\textnormal{maximize}}\limits_{\pi}\ \  \mathbb{E}_{\bm{x}, t} \left[
        {L}^{\textnormal{CLIP}}_t - c_1 L^{\textnormal{VF}}_t + c_2 S_{\pi}(\bm{s}_t)
    \right].
\end{equation}
Herein, $S_{\pi}(\bm{s}_t)$ denotes the entropy bonus to ensure sufficient exploration \cite{williams1992simple, mnih2016asynchronous, schulman2017proximal}, and $L^{\textnormal{VF}}_t$ is a squared-error loss on the estimated state value: $(V(\bm{s}_t) - V^{\textnormal{target}}(\bm{s}_t))^2$. We straightforwardly let $V^{\textnormal{target}}(\bm{s}_t) = r_t + \gamma r_{t+1} + \dots + \gamma^{T-t} r_T$. The coefficients $c_1$ and $c_2$ are pre-defined hyper-parameters.

In our implementation, we execute the aforementioned training process in Stage \uppercase\expandafter{\romannumeral2} of the 3-stage training scheme. To be specific, we optimize Eq. (\ref{eq:objective}) using an Adam optimizer \cite{kingma2014adam} with $\beta_1 = 0.9$,  $\beta_2 = 0.999$ and a learning rate of $0.0003$. We set $\gamma = 0.7$, $\epsilon = 0.2$, $c_1 = 0.5$ and $c_2 = 0.01$. The size of the mini-batch is set to 256. We train the patch proposal network $\pi$ for 15 epochs and select the model with the highest final validation accuracy, \emph{i.e.}, the accuracy when $t=T$. These hyper-parameters are selected on the validation set of ImageNet and used in all our experiments.

\subsection{Training Details for Image Classification}

\textbf{Initialization.}
As introduced in the paper, we initialize the local encoder $f_{\textsc{l}}$ using the ImageNet pre-trained models, while initialize the global encoder $f_{\textsc{g}}$ by first fine-tuning the pre-trained models with all training samples resized to $H'\!\times\! W'$. To be specific, for ResNets and DenseNets, we use the pre-trained models provided by pytorch \cite{paszke2019pytorch}, for RegNets, we use the pre-trained models provided by their paper \cite{radosavovic2020designing}, and for MobileNets-V3 and EfficientNets, we first train the networks from scratch following all the details mentioned in their papers \cite{DBLP:conf/icml/TanL19, howard2019searching} to match the reported performance, and use the obtained networks as the pre-trained models. For $H'\!\times\! W'$ fine-tuning, we use the same training hyper-parameters as the training process \cite{He_2016_CVPR, 2019arXiv160806993H, radosavovic2020designing, DBLP:conf/icml/TanL19, howard2019searching}. Notably, when MobileNets-V3 and EfficientNets are used as the backbone, we fix the parameters of the global encoder $f_{\textsc{g}}$ after initialization and do not train it any more, which we find is beneficial for the final performance of the \textit{Glance Step}. 



\textbf{Stage \uppercase\expandafter{\romannumeral1}.}
We train all networks using a SGD optimizer \cite{He_2016_CVPR, 2019arXiv160806993H, wang2020meta} with a cosine learning rate annealing technique and a Nesterov momentum of 0.9. The size of the mini-batch is set to 256, while the L2 regularization coefficient is set to 5e-5 for RegNets and 1e-4 for other networks. The initial learning rate is set to 0.1 for the classifier $f_{\textsc{c}}$. For the two encoders, the initial learning rates are set to 0.01, 0.01, 0.02, 0.005 and 0.005 for ResNets, DenseNets, RegNets, MobileNets-V3 and EfficientNets, respectively. The regularization coefficient $\lambda$ (see: Eq. (3) in the paper) is set to 1 for ResNets, DenseNets and RegNets, and 5 for MobileNets-V3 and EfficientNets. We train ResNets, DenseNets and RegNets for 60 epochs, MobileNets-V3 for 90 epochs and EfficientNets for 30 epochs.

\textbf{Stage \uppercase\expandafter{\romannumeral2}.}
We train the patch proposal network $\pi$ using an Adam optimizer \cite{kingma2014adam} with the hyper-parameters provided in Appendix \ref{app:PPO}. The standard deviation of the Gaussian distribution from which we sample the localization action $\bm{l}_{t}$ is set to 0.1 in all the experiments.

\textbf{Stage \uppercase\expandafter{\romannumeral3}.}
We use the same hyper-parameters as {Stage \uppercase\expandafter{\romannumeral1}}, except for using an initial learning rate of 0.01 for the classifier $f_{\textsc{c}}$. Moreover, we do not execute this stage for EfficientNets, since we do not witness an improvement of performance.

\begin{table*}[!t]
	\footnotesize
    \centering
    \caption{Details of the GFNets in Figure \ref{fig:main_results} of the paper}
    \vskip -0.2in
    \label{tab:gfnets}
    \setlength{\tabcolsep}{4mm}{
    \vspace{5pt}
    \renewcommand\arraystretch{1.15} 
    \begin{tabular}{l|l}
        \hline
    Backbone CNNs & GFNets \\
        \hline
    \multirow{2}{*}{ResNets}  & (1) ResNet-50, $H'=W'=96$, $T=5$ \\
    &(2) ResNet-50, $H'=W'=128$, $T=5$ \\
    \hline
    \multirow{3}{*}{DenseNets}  & (1) DenseNet-121, $H'=W'=96$, $T=5$ \\
    &(2) DenseNet-169, $H'=W'=96$, $T=5$ \\
    &(3) DenseNet-201, $H'=W'=96$, $T=5$ \\
    \hline
    \multirow{3}{*}{RegNets}  & (1) RegNet-Y-600MF, $H'=W'=96$, $T=5$ \\
    &(2) RegNet-Y-800MF, $H'=W'=96$, $T=5$ \\
    &(3) RegNet-Y-1.6GF, $H'=W'=96$, $T=5$ \\
    \hline
    \multirow{3}{*}{MobileNets-V3}  & (1) MobileNet-V3-Large (1.00), $H'=W'=96$, $T=3$ \\
    &(2) MobileNet-V3-Large (1.00), $H'=W'=128$, $T=3$ \\
    &(3) MobileNet-V3-Large (1.25), $H'=W'=128$, $T=3$ \\
    \hline
    \multirow{3}{*}{EfficientNets}  & (1) EfficientNet-B2, $H'=W'=128$, $T=4$ \\
    &(2) EfficientNet-B3, $H'=W'=128$, $T=4$ \\
    &(3) EfficientNet-B3, $H'=W'=144$, $T=4$ \\
    \hline
    \end{tabular}}
    \vskip -0.1in
\end{table*}

\textbf{Network Architecture.}
The input size ($H'$, $W'$), the maximum input sequence length $T$ and the corresponding encoders used by the GFNets in Figure \ref{fig:main_results} of the paper are summarized in Table \ref{tab:gfnets}. Note that we always let $H'\!=\!W'$.

\textbf{MS-GFNet.}
We implement MS-GFNet on top of ResNet-50 and EfficientNet-B3 with $T=4$. The glance size is set to $H'\!=\!W'\!=\!96$ and $H'\!=\!W'\!=\!128$. The focus patch size is selected among $\{96^2, 128^2, 160^2\}$ and $\{128^2, 160^2, 192^2\}$, respectively.

\subsection{Training Details for Video Recognition}
\textbf{Something-Something V1\&V2.}
In our implementation, we set $H'\!=\!W'\!=\!96$, $T=3$, and uniformly sample 8/12/16 frames from each video (corresponding to the three curves). The cascade of linear classifiers in Appendix \ref{app:recurrent_net} is used.
We always train $f_{\textsc{g}}$, $f_{\textsc{l}}$ and $f_{\textsc{c}}$ using a SGD optimizer with cosine learning rate annealing and a Nesterov momentum of 0.9. The size of the mini-batch is set to 64, while the L2 regularization coefficient is set to 1e-4. We initialize $f_{\textsc{g}}$ by fine-tuning the Something-Something pre-trained ResNet-50 using down-sampled inputs for 10 epochs with an initial learning rate of 0.1. 
Moreover, the $f_{\textsc{l}}$ is initialized by Something-Something pre-trained ResNet-50. In stage I, we train $f_{\textsc{c}}$ using randomly sampled patches for 30 epochs with an initial learning rate of 5e-3 and 0.01 for V1 and V2, respectively. Here we do not train $f_{\textsc{g}}$ and $f_{\textsc{l}}$ as we find this does not significantly improve the performance, but increases the training time. In stage II, we train $\pi$ with an Adam optimizer for 50 epochs. The same training hyper-parameters as Appendix \ref{app:PPO} are adopted. We skip stage III since we find further fine-tuning $f_{\textsc{c}}$ leads to trivial improvements.

\textbf{Jester.}
We set $H'\!=\!W'\!\in\!\{80, 96, 128\}$ (corresponding to the three curves), $T=3$, and uniformly sample 8/12/16 frames (corresponding to the three baseline dots) from each video.
All the hyper-parameters are same with Something-Something V1\&V2 dataset, unless otherwise stated. We initialize $f_{\textsc{g}}$ an initial learning rate of 0.01. In stage I, we train $f_{\textsc{c}}$ using randomly sampled patches for 10 epochs with an initial learning rate of 0.01. 

\section{Additional Results}
\begin{figure}[t]
    \begin{minipage}[t]{\linewidth}
    \centering
    \includegraphics[width=0.8\textwidth]{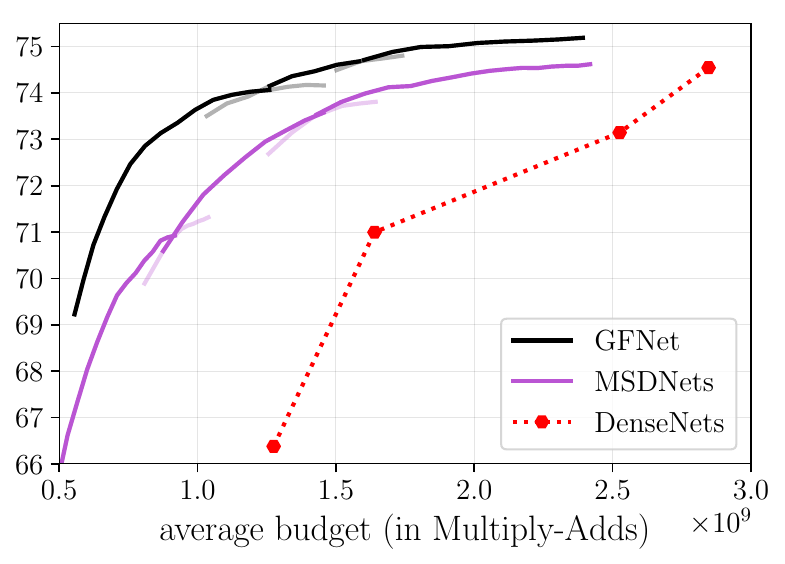}	
    \vskip -0.15in
    \caption{Performance of GFNet (based on DenseNets) versus MSDNet \cite{huang2017multi} under the \textit{Budgeted Batch Classification} setting. \label{fig:comp_dense}}  
    \end{minipage}
    \vskip -0.2in
\end{figure}

\subsection{Comparisons with MSDNet in \textit{Budgeted Batch Classification}}
The comparisons of DenseNet-based GFNets and MSDNets \cite{huang2017multi} under the \textit{Budgeted Batch Classification} setting are shown in Figure \ref{fig:comp_dense}. Following \cite{huang2017multi}, here we hold out 50,000 images from the training set as an additional validation set to estimate the confidence thresholds, and use the remaining samples to train the network (note that we use the entire training set in Figure 4 (e) of the paper). One can observe from the results that GFNet consistently outperforms MSDNet within a wide range of computational budgets. For example, when the budgets are around $1 \times 10^9$ Multiply-Adds, the test accuracy of our method is higher than MSDNet by approximately $2\%$. GFNet is shown to be a more effective adaptive inference framework than MSDNet. In addition, in terms of the flexibility of GFNet, its computational efficiency can be further improved by applying state-of-the-art CNNs as the two encoders.